%% file: main.tex
\documentclass[10pt,twocolumn,letterpaper]{article}
\usepackage[accsupp]{axessibility}
\usepackage[pagenumbers]{cvpr}              
\usepackage{multirow} 
\usepackage{makecell}
\usepackage{array}
\definecolor{tablered}{RGB}{255,200,200}
\definecolor{tableblue}{RGB}{200,220,255}
\definecolor{tablepurple}{RGB}{222,210,222}
\definecolor{tableyellow}{RGB}{255, 255, 200}

\newcommand\blfootnote[1]{%
  \begin{NoHyper}%
  \renewcommand\thefootnote{}\footnote{#1}%
  \addtocounter{footnote}{-1}%
  \end{NoHyper}%
}

\usepackage{amssymb}
\usepackage{pifont}
\usepackage{float}
\definecolor{darkgreen}{rgb}{0.2, 0.5, 0.2}


\definecolor{cvprblue}{rgb}{0.21,0.49,0.74}
\usepackage[pagebackref,breaklinks,colorlinks,allcolors=cvprblue]{hyperref}

\def\paperID{5169} 
\def\confName{CVPR}
\def\confYear{2025}

\newcommand{\dataset}{Spatial457}

\usepackage[most]{tcolorbox}
\newcommand{\hypbox}[2]{%
\begin{tcolorbox}[colback=white!98!black,colframe=white!30!black,boxsep=1.1pt,top=6.75pt]%
\vspace{1.75pt}%
\textbf{#1}\\[-0.575em]
\noindent\makebox[\textwidth]{\rule{\textwidth}{0.4pt}}
\\[0.25em]
#2
\end{tcolorbox}
}

\newtcbox{\grayround}[1][]{%
    on line, 
    colback=gray!12, 
    colframe=gray!12, 
    arc=1mm, 
    boxrule=0mm, 
    left=0.5pt, right=0.5pt, top=0.2pt, bottom=0.2pt, 
    #1 
}

\title{\dataset{}: A Diagnostic Benchmark for 6D Spatial Reasoning \\ of Large Multimodal Models}
\author{Xingrui Wang$^1$, Wufei Ma$^1$, Tiezheng Zhang$^1$, Celso M de Melo$^2$, Jieneng Chen$^1$\footnotemark[2], Alan Yuille$^1$\footnotemark[2]\\
\small $^1$Johns Hopkins University \quad $^2$DEVCOM Army Research Laboratory
}

\begin{document}
\maketitle
\input{sec/0_abstract}    
\input{sec/1_intro}
\input{sec/2_related_work}

\input{sec/3_dataset}

\input{sec/4_experiment}

\input{sec/5_real}

\input{sec/6_conclusion}

\input{sec/acknowledgement}

{
    \small
    \bibliographystyle{ieeenat_fullname}
    \bibliography{main}
}

\appendix
\input{sec/appendix/3_programs}

\input{sec/appendix/1_examples}

\input{sec/appendix/2_prompts}

\input{sec/appendix/4_vis}

\end{document}

%% file: sec/0_abstract.tex
\begin{abstract}
%
%

Although large multimodal models (LMMs) have demonstrated remarkable capabilities in visual scene interpretation and reasoning, their capacity for complex and precise 3-dimensional spatial reasoning remains uncertain. Existing benchmarks focus predominantly on 2D spatial understanding and lack a framework to comprehensively evaluate 6D spatial reasoning across varying complexities.
To address this limitation, we present \dataset{}, a scalable and unbiased synthetic dataset designed with \textbf{4} key capability for spatial reasoning: multi-object recognition, 2D location, 3D location, and 3D orientation. We develop a cascading evaluation structure,  constructing \textbf{7} question types across \textbf{5} difficulty levels that range from basic single object recognition to our new proposed complex 6D spatial reasoning tasks.
We evaluated various large multimodal models (LMMs) on \dataset{}, observing a general decline in performance as task complexity increases, particularly in 3D reasoning and 6D spatial tasks. To quantify these challenges, we introduce the Relative Performance Dropping Rate (RPDR), highlighting key weaknesses in 3D reasoning capabilities. Leveraging the unbiased attribute design of our dataset, we also uncover prediction biases across different attributes, with similar patterns observed in real-world image settings.\footnote{Distribution Statement A: Approved for public release: distribution is unlimited.}
The code and data is released in \url{https://github.com/XingruiWang/Spatial457}.

\blfootnote{$\dagger$: Equal advising.}

\end{abstract}

%% file: sec/1_intro.tex
\vspace{-0.5em}
\section{Introduction}
\vspace{-0.5em}
\label{sec:intro}

\begin{figure*}[t]
  \centering
  \includegraphics[width=0.88\linewidth]{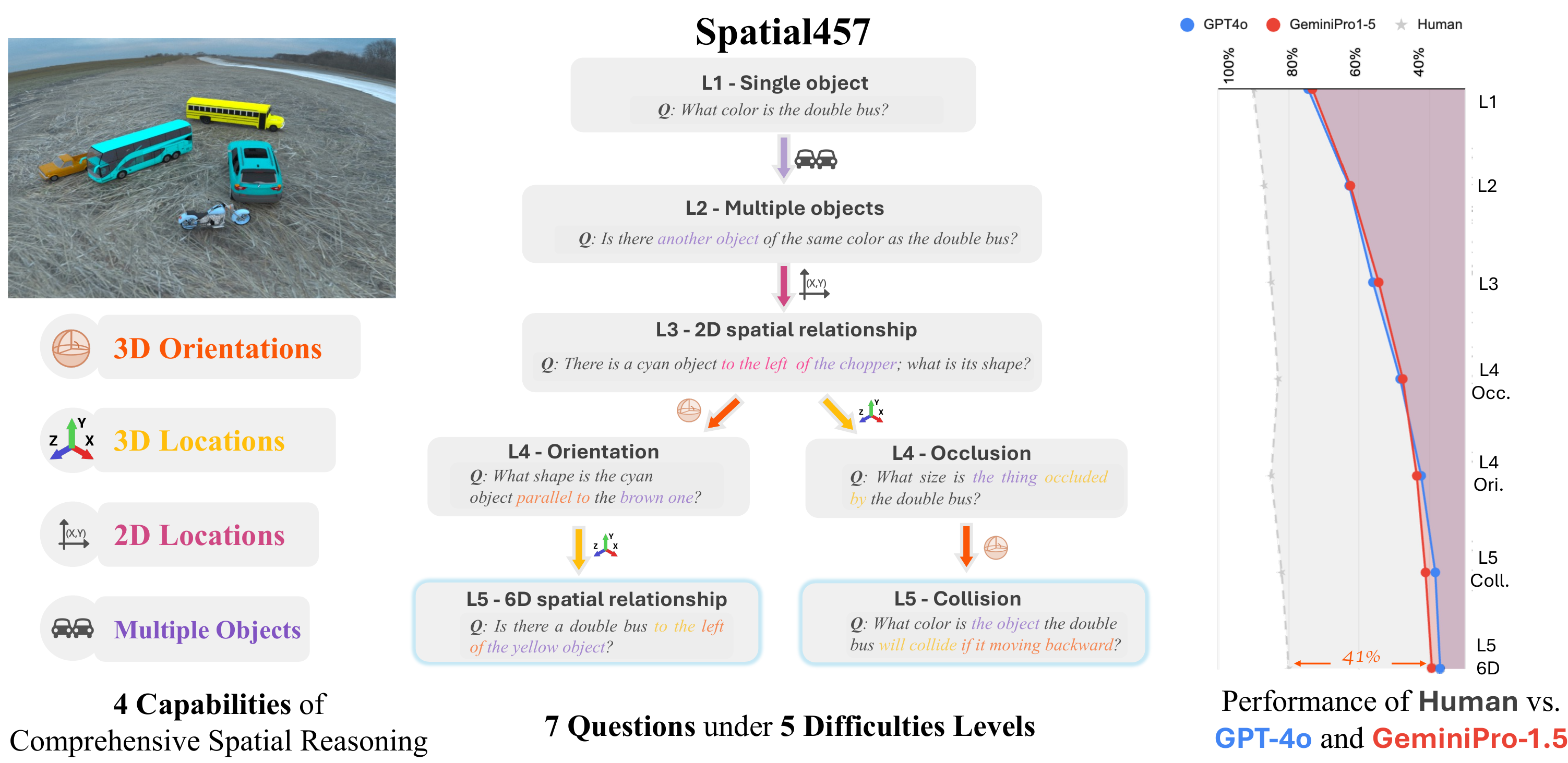}
   \caption{Overview of the \dataset{} benchmark. We define four core capabilities for spatial reasoning. By incorporating these capabilities step-by-step, the benchmark assesses models across five progressive difficulty levels and seven question types, ranging from single-object recognition to advanced 6D spatial relationships and collision prediction. Our evaluation demonstrates a significant performance drop for more advanced questions and highlights the gap between state-of-the-art (SoTA) models and human performance in complex spatial reasoning VQA tasks.}
   \label{fig:overview}
\end{figure*}

In recent years, large multimodal models (LMMs) have demonstrated remarkable reasoning capabilities, enabling them to interpret visual scenes from images and articulate their understanding through natural language~\cite{chiang2023vicuna,liu2023llava,liu2023llavanext,xu2024pllava,qwen2023vl7b,alibaba2023gemini,anthropic2023claude,intern2023vl2}. These advancements have been evaluated using a variety of benchmarks that primarily emphasize semantic features—such as object categories and appearances, spatial relationships within the 2D image plane~\cite{hu2022promptcap,goyal2017making,hudson2019gqa} (\eg left-right positioning from the camera view), and limited 2.5D features such as camera depth~\cite{tong2024cambrian}. Tasks like object detection~\cite{everingham2010pascal,kuznetsova2020open}, image captioning~\cite{kim2019dense,xu2015show,vinyals2015show} and visual question answering\cite{hudson2019gqa,yi2019clevrer,yue2024mmmu} have significantly benefited from these evaluations, driving LMMs to new heights in visual understanding.

However, a critical gap remains in assessing the ability of LMMs to comprehend and reason about the full six degrees of freedom (6D) of objects in spatial environments, which includes both 3D positions and orientations~\cite{xiang2014beyond,wang2019normalized,zhang2024omni6dpose}. Despite its significance in many real-world applications, such as robotics~\cite{an2024rgbmanip}, autonomous navigation~\cite{geiger2012we}, and augmented reality~\cite{billinghurst2015survey}, there are no available benchmarks for LMMs that evaluate 6D spatial reasoning capabilities.

\begin{figure}[!h]
  \centering
   \includegraphics[width=0.83\linewidth]{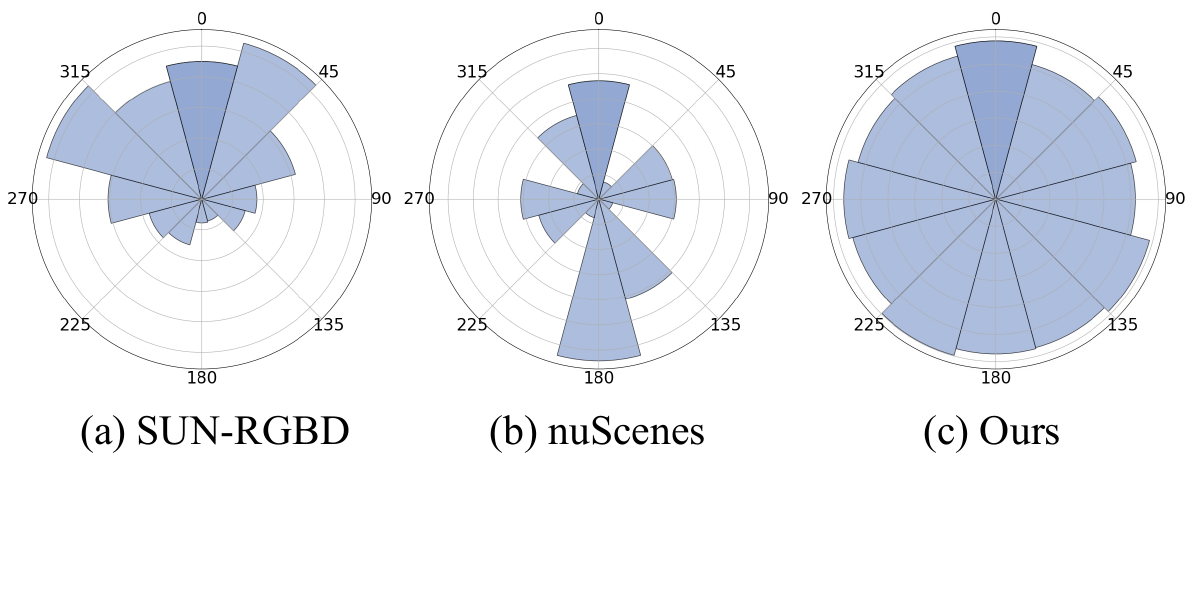}
   \caption{The distribution of 3D poses in representative 6D datasets (a\&b) highlights an \textbf{imbalance} in object orientations. This inspired us to develop a \textbf{balanced} benchmark encompassing all degrees of 3D orientation.\vspace{-1em}}
   \label{fig:sunrgbd}
\end{figure}

Building a scalable diagnostic dataset for 6D vision tasks poses considerable challenges due to the extensive annotations required across vast image collections. One strategy is to develop 6D vision question-answering benchmarks on top of existing image datasets~\cite{song2015sun,baruch2021arkitscenes,caesar2020nuscenes,brazil2023omni3d}. However, real-world image collections often exhibit significant biases in 3D locations and poses, influenced by camera perspective limitations and the prevalence of common object orientations in everyday scenes. For instance, statistics from the SUN-RGBD~\cite{song2015sun} and nuScenes~\cite{caesar2020nuscenes} datasets reveal that over 70\% of objects cluster into a single predominant direction (see \cref{fig:sunrgbd}), limiting the diversity required for comprehensive 6D reasoning evaluation.

To overcome these challenges, we introduce an unbiased and scalable synthetic dataset rendered in a realistic style, specifically designed to diagnose the 6D spatial reasoning abilities of LMMs. To comprehensively evaluate diverse aspects of 6D spatial reasoning, we focus on four core capabilities: \textbf{multi-object recognition}, \textbf{2D locations}, \textbf{3D locations}, and \textbf{3D orientation}. Multi-object recognition serves as the foundation of scene understanding, while 2D location represents the simplest way to depict the spatial relationship between objects within the image plane~\cite{johnson2017clevr,hudson2019gqa}. The 3D location factor extends understanding into three-dimensional space, capturing the depth and distance of objects, which is essential for tasks such as reasoning about occlusions~\cite{wang2023daware,chen2024spatialvlm,cheng2024spatialrgpt}. The 3D orientation introduces object rotation~\cite{wang2023daware,xiang2014beyond}, enabling models to reason about precise 3D poses of objects within a scene.

By progressively incorporating the four core capabilities, we establish five levels of difficulty, resulting in seven distinct question types. These questions span a spectrum of spatial reasoning complexity, ranging from basic semantic identification and 2D spatial relationships to advanced 3D positional understanding and orientation-based queries, culminating in comprehensive 6D spatial reasoning tasks.

Our evaluation measures performance across varied question types and examines how it is affected by different combinations of the four capabilities. This tiered design provides comprehensive benchmarks for spatial reasoning capabilities, highlighting each model's strengths and areas of vulnerability. To the best of our knowledge, this work presents the first comprehensive benchmark designed to evaluate the 6D spatial reasoning capabilities of vision-language models.

Additionally, our dataset's unbiased attribute design allows us to analyze potential biases in model predictions. By systematically exploring model performance under different attributes, we identify tendencies or limitations that may affect real-world applications. Altogether, our benchmark not only provides a foundational tool for evaluating LMMs but also serves as a guide for developing future benchmarks that support models with advanced 6D spatial reasoning.

Our main contributions are threefold:
\begin{enumerate}
    \item \textbf{An unbiased and scalable synthetic dataset}: We present a realistic, unbiased synthetic dataset designed to assess 6D spatial reasoning in LMMs. We define four core capabilities and propose five difficulty levels for spatial reasoning VQA with seven question types.
    \item \textbf{Comprehensive benchmarking of 6D capabilities}: We benchmark various LMMs across all difficulty levels. The performance variations across settings reveal potential limitations or strengths, providing valuable insights into which components should be prioritized to improve the spatial reasoning of LMMs.
    \item \textbf{Bias analysis in model predictions}: Leveraging the unbiased nature of our dataset, we conduct an in-depth bias analysis to explore how LMM predictions may be unbalanced across different attributes, especially 3D-related ones. This reveals potential weaknesses in model predictions, offering insights into limitations that may also appear in real-world scenarios.
\end{enumerate}

%% file: sec/2_related_work.tex
\section{Related Work}

\subsection{Synthetic Datasets in VQA}
\vspace{-0.5em}
The use of synthetic data in visual question answering (VQA) has gained traction as a means to circumvent the limitations and biases inherent in real-world datasets. Synthetic environments~\cite{hess2013blender, qiu2016unrealcv, sanders2016introduction} allow for controlled manipulation of scene elements, enabling researchers to probe specific reasoning abilities in VQA models without the confounding factors often present in natural imagery. Previous works~\cite{johnson2017clevr, lindstrom2022clevr, li2023super} have utilized synthetic environments to evaluate multimodal models across a range of visual reasoning tasks, typically focusing on 2D spatial relationships and object features. For instance, CLEVR~\cite{johnson2017clevr}, a synthetic dataset widely used in the VQA community, evaluates compositional reasoning by introducing artificially generated scenes with diverse object arrangements. 
Super-CLEVR-3D~\cite{wang20243d} represents a significant step forward in evaluating 3D locations, it lacks an assessment of 6D spatial relationships that consider both 3D location and 3D orientation. Our work extends this tradition by creating a synthetic dataset tailored for 6D spatial reasoning, filling a gap in current synthetic datasets that do not address the complete spatial understanding of objects in 3D space.

\vspace{-0.5em}
\subsection{LMM Benchmarks}
\vspace{-0.5em}
Existing benchmarks~\cite{goyal2017making, hudson2019gqa, singh2019towards, yue2024mmmu} for large multimodal models (LMMs) predominantly assess their performance on tasks involving 2D visual question answering and multimodal reasoning. For instance, datasets like VQAv2~\cite{goyal2017making} and GQA~\cite{hudson2019gqa} focus on answering questions based on visual content
However, these benchmarks leave a gap in evaluating models' capabilities in spatial contexts. Recent benchmarks, such as CV-Bench~\cite{tong2024cambrian}, assess spatial understanding, including depth ordering for 2.5D evaluation.
More recent work~\cite{chen2024spatialvlm,cheng2024spatialrgpt,ray2024sat} introduces questions regarding either aspect of 3D locations or rotations.
Our proposed benchmark goes beyond traditional 2D, 2.5D, and 3D location or rotation assessments by introducing tasks that challenge LMMs to understand 3D positions and orientations together. We also evaluate models in a cascading manner across five difficulty levels and summarize their performance in four reasoning abilities, along with their output biases, providing a deeper analysis of the weaknesses in spatial reasoning within multimodal AI.

%% file: sec/3_dataset.tex
\vspace{-0.3em}
\section{\dataset{}: Comprehensive Spatial Reasoning Benchmark}
\vspace{-0.3em}

To comprehensively evaluate the spatial reasoning capabilities of vision-language models, we introduce a synthetic dataset named \textbf{\dataset{}}. We define four core capabilities for comprehensive spatial reasoning (\cref{sec:4factors}) and generate scenes with full 6D annotations (\cref{sec:scenes}). The full combination of these capabilities form the new 6D spatial relationship VQA tasks (\cref{sec:6dquestions}). 

To thoroughly assess model capabilities, we define five levels of difficulty in vision question answering (VQA) tasks, converting seven question subsets that range from single-object recognition to the newly proposed 6D spatial reasoning tasks (\cref{sec:7questions}). This structured progression in question difficulty is achieved by sequentially introducing new spatial reasoning capabilities, as illustrated in \cref{fig:overview}.

\subsection{Four Capabilities for Spatial Reasoning}
\label{sec:4factors}
We define 4 capabilities for comprehensive spatial understanding in a 3D scene: \textit{multiple objects}, \textit{2D locations}, \textit{3D locations}, and \textit{3D orientations}. 


\textbf{(1) Multiple objects}. Objects are fundamental to defining scene structure~\cite{yuille2006vision}. Large vision-language models show strong results in aligning semantic features with text~\cite{radford2021learning,li2023blip}, handling multiple objects from scene becomes the first toward scene understanding. Models must recognize individual attributes and interpret relationships among them. Our dataset evaluates multi-object reasoning, such as comparison and counting. For example, in \cref{fig:overview}, the image shows two cyan objects, requiring the model to verify color similatiry and count cyan objects accurately.


\textbf{(2) 2D locations}. 2D spatial relationships are essential in prior image-based VQA tasks~\cite{johnson2017clevr,hudson2019gqa}. In traditional multimodal models~\cite{kamath2021mdetr,mao2019neuro,yi2018neural}, these relationships can be learned through object detection training~\cite{ren2016faster,carion2020end}, However, in large-scale vision-language models, explicit location information is often absent from the latent representations. Following \citet{li2023super}, we examine four spatial relationships (\texttt{left}, \texttt{right}, \texttt{front}, \texttt{back}) between objects from 2D camera views as the first taks of spatial reasoning. Questions use these relations to filter targets, such as “What is the shape of the object to the left of the chopper?” (see \cref{fig:overview}).



\textbf{(3) 3D locations}. Beyond 2D relationships, 3D locations can incorporate depth, capturing spatial relationships that 2D alone cannot represent. This is crucial for tasks involving occlusions and complex spatial arrangements. Our dataset assesses the 3D location understanding abilities of model through questions about occlusions query as in \cite{wang20243d}. For example, as shown in \cref{fig:overview}, questions may ask about an object’s position relative to others in 3D space or whether one object is occluded, challenging models to judge distances and spatial positions accurately.



\textbf{(4) 3D orientations}. The 3D orientation of objects introduces an additional layer of complexity by incorporating rotational attributes. In current computer vision tasks, 3D orientation is often derived by pose estimation methods. However, large-scale vision-language models generally lack explicit 3D orientation information in their latent representations, potentially making it challenging to accurately infer the object's orientations.
In our dataset, we evaluate models' abilities to understand 3D orientations by focusing on object poses, including their facing direction and alignments with others. For example, as depicted in \cref{fig:overview}, the model may be asked to identify whether a cyan object is facing left, oriented parallel to a brown object, or positioned in a specific spatial relationship — i.e. left, right, front, or behind— from the perspective of the objects themselves.


\vspace{-0.3em}
\subsection{Scene Rendering}
\label{sec:scenes}
\vspace{-0.3em}

We build upon the generation pipeline from prior VQA datasets \citep{johnson2017clevr,li2023super} and uniformly select 21 object models across 5 categories to construct the dataset as \citet{li2023super}. To improve the realism of our rendered dataset, we use real image style environment maps and refined object textures. For benchmarking new 6D spatial understanding ability, we add new 6D spatial relationship annotations among objects. 

\textbf{Background}. To create realistic scene backgrounds, we use HDRI images\footnote{\url{https://polyhaven.com/hdris}} with environment maps and  natural lightings. We selected 224 different outdoor scenes with normal brightness levels without high light contrast by their metadata to maintain visual consistency across images.

\textbf{Objects}. We reconstruct 3D mesh models \citep{li2023super} with realistic textures, introducing more variation and realistic appearances for objects. Similar to \citet{li2023super}, each object in our dataset is defined by distinct attributes and precise spatial locations. The original attributes include \texttt{shape}, \texttt{color}, and \texttt{size}. For benchmarking 6D spatial reasoning, each object also includes a vector specifying its 2D location \texttt{(x, y)}, 3D world coordinates \texttt{(X, Y, Z)}, and \texttt{pose} direction derived from its 3D orientation~\cite{wang20243d}.

\begin{figure}[!h]
  \centering
   \includegraphics[width=0.85\linewidth]{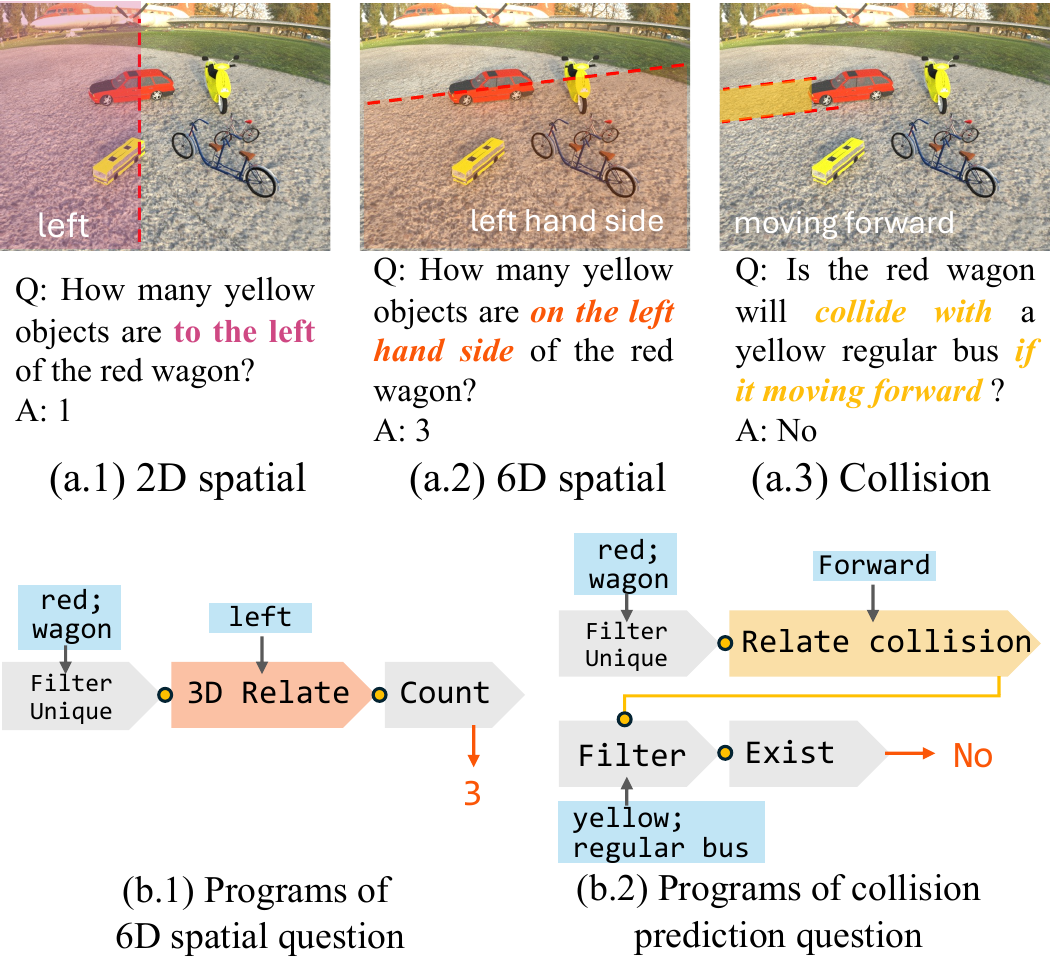}
   \vspace{-0.2em}
    \caption{ (a.1-3) Example of the 2D spatial, 3D spatial and collision with images and questions. (b.1-2) shows the new operation programs we generated for the new 6D spatial reasoning questions.}
    \vspace{-1em}
   \label{fig:programs}
\end{figure}

\textbf{Relationships}. As shown in \cref{fig:programs} (a.1 to a.3), we define three types of spatial relationships:
1. \textit{2D Spatial Relationships}: These describe the relative positioning of objects within the 2D camera view, using terms such as \texttt{left}, \texttt{right}, \texttt{front}, and \texttt{behind}.
2. \textit{6D Spatial Relationships}: Extending spatial descriptions into 3D space, these relationships incorporate both the 3D position and orientation of objects to define relative positions from a target object's perspective, including \texttt{left}, \texttt{right}, \texttt{front}, and \texttt{behind}.
3. \textit{Collision Relationships}: This relationship describes potential collisions if the target object moves \texttt{forward} or \texttt{backward}. Collision prediction involves assessing the orientation and 3D location of target objects relative to others to determine possible collisions.



\vspace{-0.3em}
\subsection{New Questions for 6D Spatial Reasoning}
\label{sec:6dquestions}
\vspace{-0.3em}

Based on the four capabilities defined in \cref{sec:4factors}, we construct new question types for \textbf{6D spatial reasoning} and \textbf{collision prediction}. These questions require an understanding of both 3D location and 3D orientation across multiple objects to generate accurate answers. This level of complexity represents the L5 difficulty level shown in \cref{fig:overview}.

The 6D spatial relationship questions challenge the model to understand the spatial relationship involving both 3D position and orientation from a target object's perspective (shown in \cref{fig:programs} (a.2)). Compared with prior VQA benchmarks focusing on 2D~\cite{johnson2017clevr,li2023super}, we introduce new 3D spatial reasoning operations, \texttt{3D Relate($\cdot$)}, to determine relative positions like \texttt{left}, \texttt{right}, \texttt{front}, and \texttt{behind} within a 3D coordinate space (see \cref{fig:programs} (b.1)). For example, a question might ask, “How many objects are to the right of the red wagon?” requiring the model to filter and count objects based on their orientation and relative 3D position, which would yield different answers compared to similar 2D left and right questions (\cref{fig:programs} (a.1)). This task is particularly challenging as it requires models to integrate 3D spatial relationships while managing occlusions and depth perception—capabilities not typically covered in simpler 2D spatial tasks.

The collision prediction questions, as shown in \cref{fig:programs} (a.3), extend 6D spatial reasoning by introducing motion-based inference. These questions assess the model’s ability to anticipate future states and potential interactions (collisions) based on accurate estimation of 3D location and direction, denoted as \texttt{forward} or \texttt{backward}. For instance, a question might ask, ``Will the red wagon collide with a yellow bus if it moves forward?”. This requires the model to consider the orientation and direction of movement relative to surrounding objects to predict possible outcomes. Such tasks test the model's advanced spatial reasoning about 3D orientation and position, and its capacity to predict accurate trajectories and directions within the future scene.

\subsection{Questions under Other 4 Difficulty Levels}
\label{sec:7questions}

The L5 questions above represent the most challenging cases for 6D spatial reasoning, testing all four capabilities in \cref{sec:4factors}. However, the purpose of this benchmark is to comprehensively diagnose each factor in spatial reasoning. Thus, we design a structured roadmap to evaluate the model's performance across different levels of spatial reasoning, assessing its capabilities in each specific factor.

By composing the four factors of comprehensive spatial reasoning, we define an additional four levels of question difficulty based on the generated scenes, encompassing a total of seven tasks in our benchmark. As the complexity of spatial reasoning increases with the introduction of additional factors, we organize these questions into a structured roadmap, as illustrated in \cref{fig:overview}. Although many questions are reimplementations from previous benchmarks~\cite{li2023super,wang20243d} on our new images, we aim to integrate these into a unified benchmark to provide deeper insights into the model's spatial reasoning capabilities. The composition of the four capabilities at each level, along with comparisons to previous VQA benchmarks, is shown in \cref{tab:levels}.


\textbf{Level 1 - single object questions}~\cite{johnson2017clevr,li2023super}. We begin with the questions with single objects. These questions are generated using a filtered set of templates from \citet{li2023super}. We select five templates involving only one object in the question, such as ``What color is the double bus?" in \cref{fig:overview}.
This question type provides a base score for the tested vision-language model. The comparison with L1 helps to determine whether the decrease of performance are due to challange of spatial reasoning or object recognition.

\textbf{Level 2 - multiple objects questions}~\cite{johnson2017clevr,li2023super}. Level 2 questions extend Level 1 by involving multiple objects. These questions involves understand the relationship between objects and counting. For example, ``Is there another object of the same color as the double bus?” as shown in \cref{fig:overview}, where the question templates follows \citet{li2023super}.
We evaluate the model's capacity for reasoning in scenes with multiple entities as a critical step toward understanding spatial relationships in next levels.
 
\textbf{Level 3 - 2D spatial questions}~\cite{johnson2017clevr,li2023super}. Level 3 questions further introduce 2D spatial relationships among multiple objects, requiring the model to understand the position of objects and their relationships from the 2D camera's perspective. For example, “What shape is the object to the left of the chopper?” as depicted in \cref{fig:overview}.
The 2D relationship is the first step of spatial reasoning.

\textbf{Level 4 - 3D pose questions}~\cite{wang20243d}. Level 4 questions extend 3D orientation reasoning beyond the 2D spatial relationships in Level 3. Beyond appearance attributes (\texttt{shape}, \texttt{color}, \texttt{size}) or 2D locations, these questions require understanding 3D orientation, including the rotation angle or facing direction of an object from the camera's perspective. For example, “What shape is the object that is parallel to the brown object?” as shown in \cref{fig:overview}. 3D pose questions test the model's ability to interpret scenes with 3D awareness, involving rotation and alignment with other objects.

\textbf{Level 4 - occlusion questions}~\cite{wang20243d}. In addition to 3D orientation, incorporating 3D location presents another challenge beyond the Level 3 questions on 2D spatial relationships. A unique aspect of 3D spatial reasoning from images is occlusion. The occlusion questions in Level 4 require the model to comprehend depth perception and the relative positions of objects in 3D space, determining which object is closer to the camera within overlapping areas. For example, an occlusion question might ask, “Is the red wagon occluded by the yellow bus?” as shown in \cref{fig:overview}. This level evaluates the model's ability to reason about occlusions and depth perception, which are essential for understanding complex spatial relationships in 3D scenes.
\vspace{-1em}

\begin{table}[!h]
\centering
\renewcommand{\arraystretch}{1.5}
\resizebox{1.00\columnwidth}{!}{
\begin{tabular}{c|cccc|c}
\makecell{Question\\ Level}  & \rotatebox{90}{Multi-object}& \rotatebox{90}{2D location} & \rotatebox{90}{3D location} & \rotatebox{90}{3D Orientation} & VQA Benchmarks \\
\midrule
L1 &  &  & & & spatial-agnostic bench (e.g., TextVQA~\cite{singh2019towards}) \\
L2 & $\checkmark$ &  &  & & CLEVR-Math~\cite{lindstrom2022clevr}, CLEVR-Count~\cite{johnson2017clevr} \\
L3 & $\checkmark$ & $\checkmark$ & & &  VQA~\cite{goyal2017making}, GQA~\cite{hudson2019gqa}, CLEVR~\cite{johnson2017clevr}, Super-CLEVR~\cite{li2023super} \\
\hline
L4 (a)  & $\checkmark$ & $\checkmark$ & $\checkmark$ & &  CV-Bench (DepthOrder)~\cite{tong2024cambrian}, Super-CLEVR-3D~\cite{wang20243d}  \\
L4 (b) &  $\checkmark$ & $\checkmark$ & & $\checkmark$ & Super-CLEVR-3D~\cite{wang20243d}  \\
\hline
L5 & $\checkmark$ & $\checkmark$ & $\checkmark$ & $\checkmark$ & \textbf{\dataset{}}   \\
\end{tabular}
}
\caption{Comparison of VQA Benchmarks by Factors and Question Levels. Compared with existing VQA benchmarks, \dataset{} includes all four spatial reasoning factors, providing a comprehensive evaluation of 6D spatial understanding.}
\vspace{-0.8em}
\label{tab:levels}
\end{table}

\vspace{-0.7em}
\subsection{Evaluation Protocols}
\label{sec:RPDR}
\vspace{-0.3em}
To better evaluate the model's spatial reasoning capabilities, we propose a structured evaluation protocol that measures the relative capability of models on each factor for spatial reasoning. We define this evaluation metric as the Relative Performance Dropping Rate (RPDR), which assesses the decline in performance after introducing a new factor in the questions. Based on the roadmap in \cref{fig:overview}, the RPDR for each factor is defined as follows:
\vspace{-0.5em}
{\small
\begin{align*}
    \text{RPDR}_{\text{Multi-Obj.}} &= Acc_{\text{Level 1}} \Rightarrow Acc_{\text{Level 2}} \\ 
    \text{RPDR}_{\text{2D Loc.}} &= Acc_{\text{Level 2}} \Rightarrow Acc_{\text{Level 3}} \\
    \text{RPDR}_{\text{3D Loc.}} &= \text{Avg} \left( 
    \begin{array}{c}
        Acc_{\text{Level 3}} \Rightarrow Acc_{\text{Level 4 (Occ.)}},\\
        Acc_{\text{Level 4 (3D Pose)}} \Rightarrow Acc_{\text{Level 5 (6D)}}
    \end{array}
    \right) \\
   \text{RPDR}_{\text{3D Ori.}} &= \text{Avg} \left( 
    \begin{array}{c}
        Acc_{\text{Level 3}} \Rightarrow Acc_{\text{Level 4 (3D Pose)}},\\
        Acc_{\text{Level 4 (Occ.)}} \Rightarrow Acc_{\text{Level 5 (Col.)}}
    \end{array}
    \right)
\end{align*}}
where $Acc_{\text{Level}}$ denotes the model's accuracy on the corresponding questions and $(Acc_{\text{Q1}} \Rightarrow Acc_{\text{Q2}})=\frac{Acc_{\text{Q2}}}{Acc_{\text{Q1}}}$ denotes the relative performance dropping rate from $Q1$ to $Q2$. The results of RPDR will be discussed in \cref{sec:RPDR_results}.

%% file: sec/4_experiment.tex
\begin{table*}[!t]
\centering
\small
\begin{tabular}{l|ccccccc}
\toprule
\textbf{Difficulty Level} & \multicolumn{1}{c}{Level 1} & \multicolumn{1}{c}{Level 2} & \multicolumn{1}{c}{Level 3} & \multicolumn{2}{c}{Level 4} & \multicolumn{2}{c}{Level 5} \\
\midrule
\textbf{Questions type} & Single Object & Multi-Obj. & 2D Spatial & Occlusion & 3D Pose & Collisions & 6D Spatial \\
\midrule
Random & 33.05 & 32.77 & 33.47 & 22.04 & 18.99 & 21.02 & 19.41 \\
\midrule
GPT-4o~\cite{openai2023gpt4} & \cellcolor{tablered}74.46 & \cellcolor{tablered}62.88 & \underline{56.14} & \cellcolor{tablered}48.40 & \underline{42.41} & \underline{38.41} & \underline{37.01} \\
GeminiPro 1.5~\cite{team2024gemini} & \underline{73.26} & \underline{62.54} & 54.49 & \underline{47.65} & \cellcolor{tablered}43.67& \cellcolor{tablered}41.19 & \cellcolor{tablered}39.36 \\
Claude 3.5 Sonnet~\cite{anthropic2023claude} & 68.24 & 57.40 & 54.19 & 30.84 & 38.40 & 35.34 & 33.48 \\
\midrule
Qwen2-VL-7B-Inst.~\cite{qwen2023vl7b} & \cellcolor{tableblue}71.96 & \cellcolor{tableblue}61.44 & 55.34 & 27.87 & \cellcolor{tableblue}34.29 & \cellcolor{tableblue}36.58 & 33.75 \\
InternVL2-8B~\cite{intern2023vl2} & 58.11 & 58.76 & \cellcolor{tablered} 57.40 & \cellcolor{tableblue}33.25 & 32.32 & 34.30 & \cellcolor{tableblue}34.30 \\
LLaVA-v1.5-7B~\cite{liu2023llava} & 44.87 & 44.72 & 42.20 & 24.34 & 24.55 & 23.63 & 23.86 \\
LLaVA-NeXT-vicuna-7B~\cite{liu2023llavanext} & 50.72 & 49.47 & 46.01 & 29.68 & 29.35 & 31.95 & 31.95 \\
LLaVA-NeXT-llama3-8B~\cite{touvron2023llama} & 52.15 & 49.73 & 45.92 & 30.31 & 29.77 & 32.12 & 32.12 \\
\midrule
PO3D-VQA~\cite{wang20243d} & 86.46 & 82.55 & 80.64 & 70.49 & 81.40 & 68.12 & 71.06 \\
Human & 89.97 & 86.83 & 84.95 & 82.76 & 84.95 & 81.82 & 79.94 \\
\bottomrule
\end{tabular}
\caption{Performance comparison across all 7 question types under 5 difficulty levels. As task complexity increases from single-object questions to 6D spatial relationships, performance generally decreases in most cases, highlighting the challenges in handling multi-object interactions, 3D orientations, and predictive spatial reasoning. The highest score among all models for each question type is marked in \colorbox{tablered}{red}, and the second-highest scores are \underline{underlined}. The best performance among open-source models is marked in \colorbox{tableblue}{blue}. PO3D-VQA is the state-of-the-art (SOTA) 3D-aware neural symbolic reasoning model as proposed in ~\cite{wang20243d}, and is listed as the potential upper bound of evaluated methods. The comparison with human performance also shows an increasing gap for LLMs as task difficulty increases.}
\label{tab:performance}
\end{table*}

\vspace{-0.4em}
\section{Experiments}
\vspace{-0.3em}

\subsection{Experimental Setup}
\vspace{-0.3em}
We generate 1,000 images to test the performance of large vision-language models. For the seven types of questions, we generate 670 single-object questions (\textit{L1-Single}), 4,995 multiple-object questions (\textit{L2-Multi-obj.}), 4,995 2D spatial relationship questions (\textit{L3-2D-Spatial}), 3,534 occlusion questions (\textit{L4-Occlusion}), 4,555 3D pose questions (\textit{L4-3D-Pose}), 3,166 collision questions (\textit{L5-Collision}), and 1,837 6D spatial relationship questions (\textit{L5-6D-Spatial})

We evaluate two series of large vision-language models on \dataset{}: (1) API-based large language models, including GPT-4o~\cite{openai2023gpt4}, Gemini-1.5-Pro~\cite{team2024gemini}, and Claude-3.5-Sonnet~\cite{anthropic2023claude}; and (2) open-source large vision-language models, including Qwen2-VL-7B-Instruct~\cite{qwen2023vl7b}, InternVL2-8B~\cite{intern2023vl2}, llava-v1.5-7b~\cite{liu2023llava}, llava-next-vicuna-7b~\cite{liu2023llavanext}, and llava-next-llama3-8b~\cite{touvron2023llama}. The evaluation follows the format of Vlmevalkit~\cite{duan2024vlmevalkit}.
The details of the model implementation including prompts are listed in Appendix. 

\vspace{-0.3em}
\subsection{Evaluation Results}
\vspace{-0.3em}
In \cref{tab:performance}, we present the performance of various large vision-language models across different levels of difficulty on \dataset{}. The levels range from basic single-object tasks to complex 6D spatial reasoning, with increasing difficulty as more factors are introduced. We compare both API-based models (e.g., GPT-4o, GeminiPro1.5, Claude3-5V) and open-source models (e.g., Qwen2-VL-7B-Instruct, InternVL2-8B, LLaVA series) to assess their spatial reasoning capabilities.

\textbf{Comparison across difficulty levels.} We evaluate model performance across different levels of difficulty in \dataset{}, ranging from basic single-object tasks to complex 6D spatial reasoning. As shown in \cref{tab:performance}, performance generally decreases as the complexity of spatial relationships increases, highlighting the challenges models face when handling multi-object interactions, 3D orientations, and predictive spatial reasoning tasks.

\textbf{Comparison between API models and open-source models.} From \cref{tab:performance}, we observe that API models generally outperform open-source models across all difficulty levels. In 2D spatial relationship questions, InternVL2-8B achieves slightly higher results than the API models, while GPT-4o is only 1\% lower. Qwen2-VL-7B-Instruct achieves comparable results, especially in simpler tasks like single-object and multiple-object reasoning. This analysis highlights the strengths of API models in complex spatial reasoning tasks.

\textbf{Comparison among API models.} API-based models, \ie GPT-4o, GeminiPro1.5, and Claude3-5V, demonstrate strengths across all difficulty levels. For instance, GPT-4o excels in 2D and 3D spatial tasks, while GeminiPro1.5 shows robust performance in collision prediction tasks. This comparison provides insights into the unique capabilities and limitations of each API model in handling spatial reasoning across different levels of complexity.

\vspace{-0.2em}
\subsection{RPDR Analysis for 4 Capabilities}
\label{sec:RPDR_results}
\vspace{-0.2em}


As outlined in \cref{sec:RPDR}, we use the Relative Performance Dropping Rate (RPDR) to analyze model performance across the four core factors of \dataset{}: multi-object reasoning, 2D location understanding, 3D location comprehension, and 3D orientation interpretation. The RPDR analysis provides a structured approach to quantify the decline in accuracy with the introduction of each new factor, as detailed in \cref{tab:spatial_performance}. Our findings reveal that all models exhibit weaknesses in 3D reasoning. Specifically, GPT-4o and GeminiPro-1.5 perform the worst in 3D orientation tasks, while the other two struggle most with 3D location.

\begin{table}[!h]
\centering
\footnotesize
\begin{tabular}{l|cccc}
\toprule
\textbf{Model} & \textbf{Multi.} & \textbf{2D Loc.} & \textbf{3D Ori.} & \textbf{3D Loc.} \\
\midrule
GPT-4o~\cite{openai2023gpt4} & 84.45 & 89.27 & \cellcolor{tableyellow}77.46 & 86.74 \\
GeminiPro 1.5~\cite{team2024gemini} & 85.37 & 87.13 & \cellcolor{tableyellow}83.28 & 88.79 \\
Claude 3.5 S.~\cite{anthropic2023claude} & 84.12 & 94.41 & 85.43 & \cellcolor{tableyellow}72.05 \\
Qwen2-VL-7B~\cite{qwen2023vl7b} & 85.39 & 90.06 & 80.99 & \cellcolor{tableyellow}74.40 \\
\bottomrule
\end{tabular}
\caption{RPDR (\%) analysis on the four spatial reasoning capabilities, where higher values indicate better performance. For each row, we \colorbox{tableyellow}{highlight} the lowest RPDR score among all factors.}
\vspace{-1em}

\label{tab:spatial_performance}
\end{table}

\vspace{-0.5em}
\subsection{Bias Analysis}
\vspace{-0.5em}
In this section, we analyze prediction biases in large language models (LLMs) regarding their understanding of key attributes, specifically shape, color, size, and pose. Leveraging the unbiased generation of dataset images and questions in \dataset{}, we examine how these models may demonstrate attribute-specific biases in their predictions.

\begin{figure}[h]
  \centering
   \includegraphics[width=0.9\linewidth]{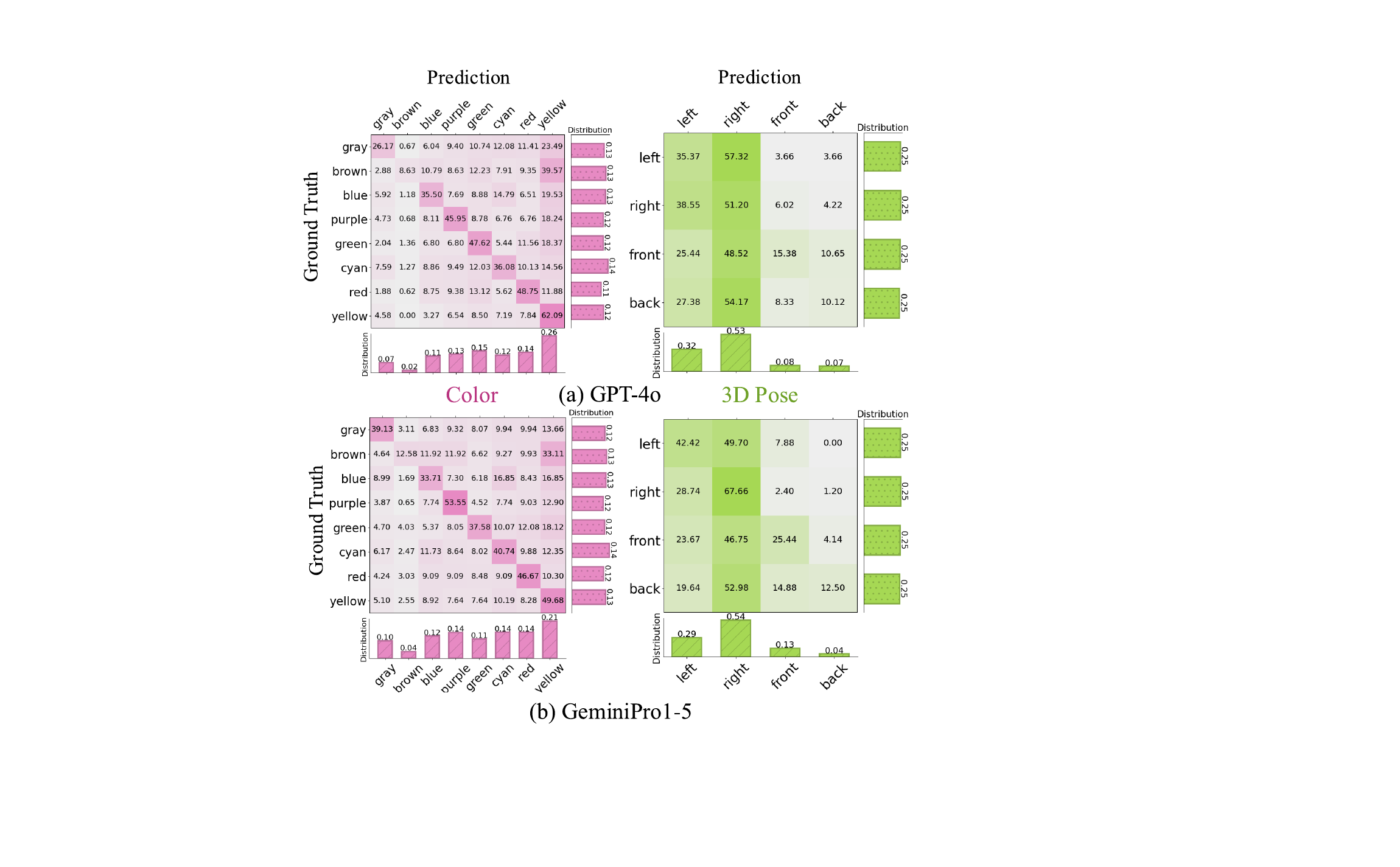}
\caption{The distribution of \texttt{color} and \texttt{pose} attributes in the L4-3D-Pose task for GPT-4o and Gemini-Pro1.5. Although the ground truth labels are well balanced, the predicted colors and poses for both models are imbalanced, with ``yellow" being the most frequent color and ``front" the most frequent pose.}
   \label{fig:exp_bias}
\end{figure}

We first compute the confusion matrices and distribution statistics for both ground truth and model predictions across each attribute (shape, color, size, and pose). For example, the confusion matrix plots for color and pose attributes of GPT-4o and GeminiPro1.5, together with the distribution of prediction and ground truth label are shown in ~\cref{fig:exp_bias}. We find both GPT-4o and GeminiPro1.5 exhibit bias in both color and pose predictions. These results highlight areas for improvement to reduce bias and enhance predictive reliability across models.

To quantify prediction variability, we calculate the Coefficient of Variation (CV) for each attribute as follows:
{\small
\begin{equation}
    \text{CV} = \frac{\text{Std}(p_1,\ldots,p_k)}{\text{Mean}(p_1,\ldots,p_k)},\vspace{-0.5em}
\end{equation}}
where $\text{Std}(\cdot)$ and $\text{Mean}(\cdot)$ are the standard deviation and mean of the probability distribution $p_i$ for each attribute label $i$ (e.g., left," "right" for pose). 

\cref{tab:exp_bias} shows the CV values for shape, color, size, and pose attributes across different models. Lower CV values indicate more consistent predictions with less bias, which reveals significant biases in model predictions across attributes. Higher CV values, especially in color and pose, indicate consistency challenges. 

\begin{table}[h]
\footnotesize
\centering
\begin{tabular}{lccccc}
\toprule
\textbf{Models} & $Avg.\downarrow$ & Shape & Color & Size & Pose\\
\midrule
GPT4o~\cite{openai2023gpt4} & \cellcolor{tablered}0.485 & 0.623 & 0.521 & \cellcolor{tablered}{0.043} & 0.752 \\
GeminiPro1.5~\cite{team2024gemini} & 0.498 & \cellcolor{tablered}{0.567} &  \cellcolor{tablered}{0.366} & 0.297 & 0.760 \\
Claude3.5V-S~\cite{anthropic2023claude} & 0.720 & 0.777 & 0.465 & 0.574 & 1.063 \\
Qwen2-VL-7B~\cite{qwen2023vl7b} & 0.545 & 0.868 & 0.558 & 0.067 & \cellcolor{tablered}{0.688} \\
\midrule
Human & 0.093 & 0.110 & 0.082 & 0.055 & 0.126 \\
\bottomrule
\end{tabular}
\caption{Coefficient of Variation (CV) for predictions on each attribute in the \textit{L4-3D-Pose} task. We evaluate three API models and one open-source model, calculating the CV across four attributes (shape, color, size, 3D pose) based on the distribution of predicted labels. Lower CV values indicate less bias. The first column shows the average CV across attributes, and GPT4-o has the lowest bias.}
\vspace{-1em}
\label{tab:exp_bias}
\end{table}

\vspace{-0.4em}
\subsection{Comparison to the 3D Symbolic Model}
\vspace{-0.4em}

As neural symbolic models achieve state-of-the-art (SOTA) performance on many prior synthetic datasets, we implement the 3D-aware neural symbolic model PO3D-VQA~\cite{wang20243d} on our new image domain and all levels of questions. We train PO3D-VQA on a separate set of 20,000 images and questions and report its performance on the same test dataset in \cref{tab:performance}. 
A direct comparison between the 3D neural symbolic model and large language models is not entirely fair, as each module of the neural symbolic model (6D pose estimation, attribute classification, and language reasoning) is trained independently as a sub-task. However, we include this model to illustrate the upper bound that a data-driven machine learning model can achieve on our dataset.

\vspace{-0.5em}
\subsection{Qualitative Results}
\vspace{-0.5em}

\begin{figure}[h]
  \centering
   \includegraphics[width=0.9\linewidth]{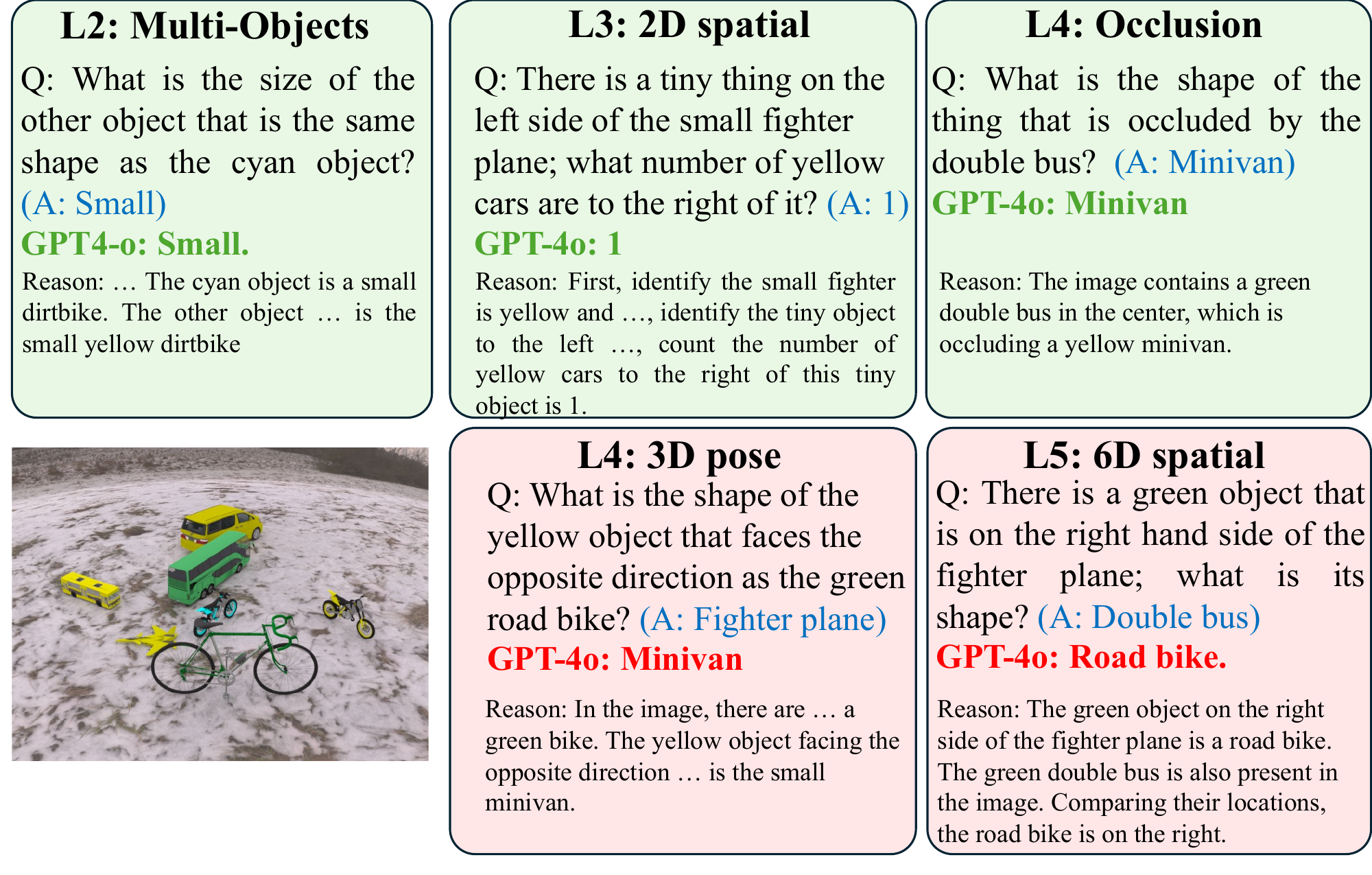}
   \vspace{-0.5em}
\caption{Example of GPT-4o across different spatial reasoning tasks on the same image with the ground truth answer marked in blue. GPT-4o answers multi-object, 2D spatial, and occlusion tasks correctly (in the green block) but fails on 3D pose and 6D spatial questions (the red block). }
\label{fig:exp_vis}
\end{figure}

\cref{fig:exp_vis} presents an example of GPT-4o's performance across various spatial reasoning tasks on the same image. GPT-4o correctly answers multi-object, 2D spatial, and occlusion tasks but fails on 3D pose and 6D spatial questions, indicating challenges in 3D orientation understanding in this case. As we prompt the models to output the answer as well as the reasoning steps, the models correctly detect the existence of objects (e.g. in \cref{fig:exp_vis} the yellow minivan in L4-Pose or the fighter plane and road bike in L5 questions, but has the wrong understand to the orientations or 3D locations.
\vspace{-0.3em}
\vspace{-0.5em}
\vspace{-0.3em}

%% file: sec/5_real.tex
\vspace{-1em}
\section{Extension to Real Dataset}
\vspace{-0.5em}
\label{sec:real}
We extend the setting of 3D questions to real-world image data. Due to limitations in fully annotated attributes such as category, color, 3D orientation, position, and occlusion, we focus on questions from the \textit{L4-Pose} category. Using the scene structure from SUN-RGBD~\cite{song2015sun}, we construct a new subset, \textit{L4-Pose-Real}, which involves only the category and pose attributes. To reduce ambiguity in object localization, we add a red bounding box around target objects, ensuring clarity in identifying categories within the scene.

\begin{figure}[h]
  \centering
   \includegraphics[width=0.9\linewidth]{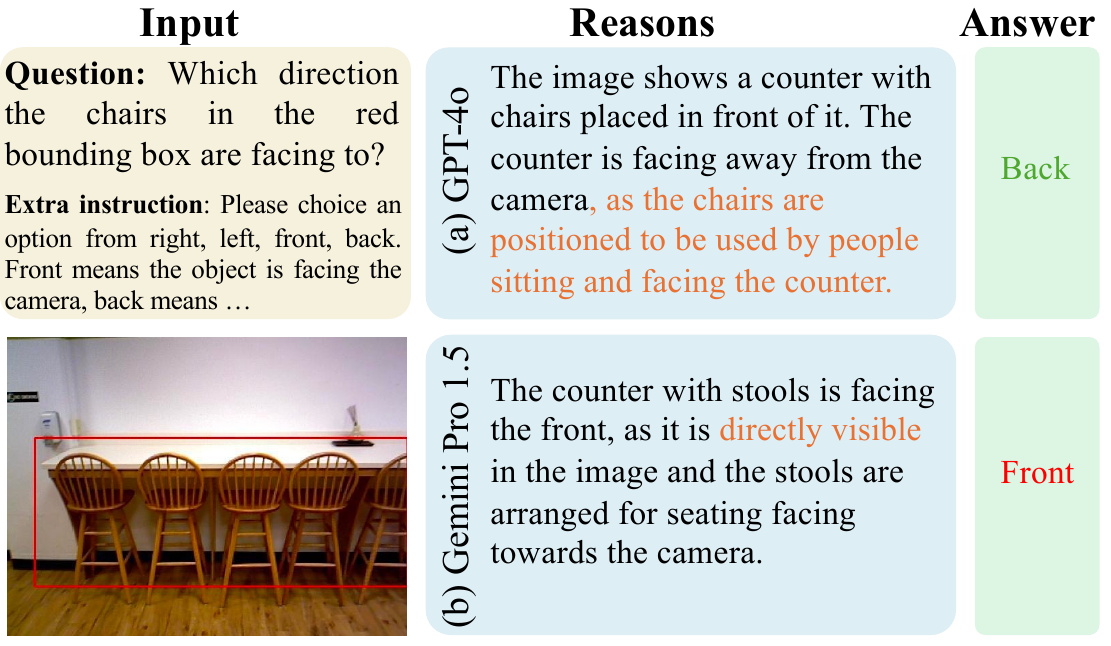}
   \vspace{-0.3em}
    \caption{Example of real-world 3D pose reasoning tasks \textit{L4-Pose-Real}. Here we shows the reasoning and answers provided by LLMs where  GPT-4o is correct and Gemini-Pro 1.5 is wrong.  But the reasoning process reveals these two LLMs are both reasoning 3D orientation from the command sense or 2D visibility.}
    \vspace{-1em}
   \label{fig:exp_real}
\end{figure}

We provide the question along with the image as input to the LLM. In this example in \cref{fig:exp_real}, GPT-4o provides the correct answer, while Gemini-Pro 1.5 is incorrect. However, the reasoning process reveals that both models rely on common sense knowledge or 2D visibility cues to interpret the 3D orientation.

\begin{figure}[h]
  \centering
   \includegraphics[width=0.9\linewidth]{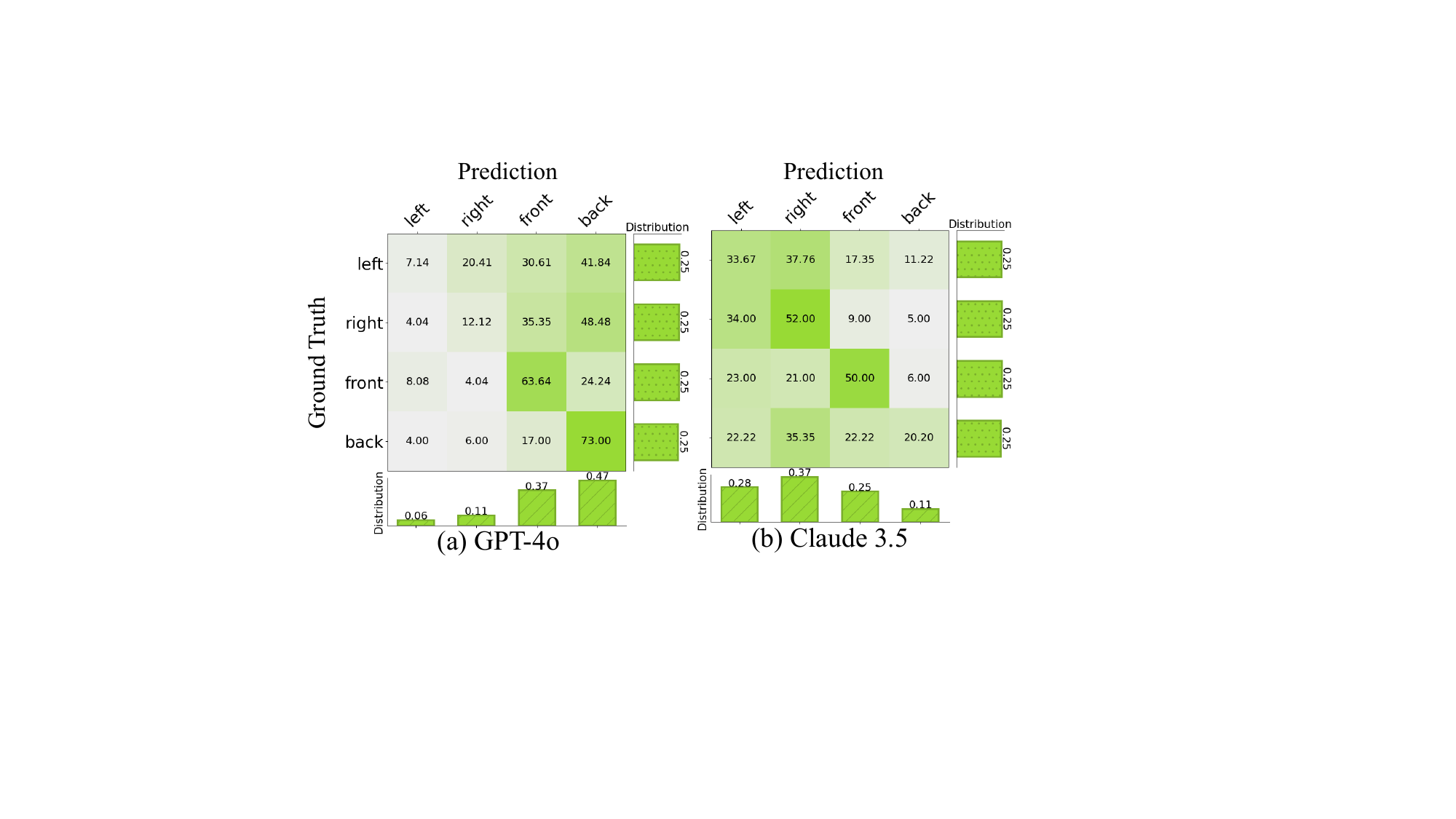}
   \caption{Distribution of 3D poses for in \textit{L4-Pose-Real} task.}
   \vspace{-1em}
   \label{fig:exp_bias_real}
\end{figure}

As shown in ~\cref{tab:exp_real}, Gemini Pro 1.5 achieves the highest accuracy (40.50\%) in \textit{L4-Pose-Real}, and the gap between the three API-based models and the open-source model Qwen2-VL-7B-Instruct (26.50\%) is larger. However, all these models perform significantly lower than humans, which have accuracy 92.64\%.  We further analyze the bias of predictions in this real-world question set. Among the models, Claude3-5V Sonnet exhibits the lowest CV (0.374), indicating relatively stable predictions, while Qwen2-VL-7B-Instruct has the highest CV (0.934). The distribution of ground truth and predicted pose and the confusion matrix are shown in \cref{fig:exp_bias_real}.

\begin{table}[ht]
\centering
\small
\begin{tabular}{lcc}
\toprule
\textbf{Models} & \textbf{Accuracy (\%)}  $\uparrow$  & \textbf{CV} $\downarrow$ \\
\midrule
GPT-4o~\cite{openai2023gpt4} & 38.75 & 0.690 \\
GeminiPro1-5~\cite{team2024gemini} & \cellcolor{tablered}40.50 & 0.762 \\
Claude3-5V Sonnet~\cite{anthropic2023claude}  & 38.75 & \cellcolor{tablered}0.374 \\
Qwen2-VL-7B-Instruct~\cite{qwen2023vl7b}  & 26.50 & 0.934 \\
\midrule
Human & 92.64 & 0.197 \\
\bottomrule
\end{tabular}
\vspace{-0.9em}
\caption{Accuracy and Coefficient of Variation (CV) for \textit{L4-Pose-Real} task: Comparison of accuracy and CV for GPT-4o, GeminiPro1.5, Claude3-5V, and Qwen2-VL-7B-Instruct on real-world 3D pose tasks. Human performance is included as a reference.}
\vspace{-1em}
\label{tab:exp_real}
\end{table}

%% file: sec/6_conclusion.tex
\vspace{-0.6em}
\section{Conclusion}
    \vspace{-0.6em}
In this work, we introduce \dataset{}, a scalable and unbiased benchmark designed to comprehensively assess 6D spatial reasoning capabilities in LMMs. \dataset{} extends traditional 2D and 3D VQA benchmarks by incorporating complex 6D spatial reasoning tasks, structured across five levels of difficulty. Through evaluations, we observe that current LMMs demonstrate strong performance in basic object recognition and 2D spatial relationships, but struggle significantly with 3D spatial understanding and collision prediction tasks. Additionally, our analysis of prediction biases uncovers attribute-specific inconsistencies that could affect model reliability in real-world applications. 
\textbf{Limitations.} To real images settings, current evaluation limites to L3 pose questions. We are actively working on building the additional required annotations for these real images to expand the domain of the proposed benchmark.

%% file: sec/acknowledgement.tex
\section*{Acknowledgement}
This work is supported by ONR with award N00014-23-1-2641, from ARL Army Research Laboratory with award W911NF2320008.

%% file: sec/appendix/3_programs.tex
\clearpage
\setcounter{page}{1}
\setcounter{section}{0}
\renewcommand{\thesection}{\Alph{section}}
\maketitlesupplementary

\section{New programs for 6D spatial reasoning}

As introduced in \cref{sec:6dquestions}, we create two types of new questions: \textbf{6D spatial reasoning} and \textbf{collision prediction}. 

The 6D spatial reasoning questions introduce the new program \grayround{\texttt{3D Relate}}, an extension of \grayround{\texttt{Relate}} from \citet{johnson2017clevr}. It takes two arguments: a direction (\texttt{left}, \texttt{right}, \texttt{front}, or \texttt{behind}) and an object ID specifying the target.  Unlike \grayround{\texttt{Relate}}, which operates in a 2D plane from the camera's view, \grayround{\texttt{3D Relate}} works in 3D space from the object's perspective, as described in \cref{sec:6dquestions}.

For collision prediction, we introduce the new program \grayround{\texttt{Relate Collision}}. This operation expands on the spatial reasoning of \grayround{\texttt{3D Relate}} by incorporating motion-based inference. It assesses whether the movement of a target object, specified with directions like \texttt{forward} or \texttt{backward}, will result in a collision with other objects in the scene. This program enables models to predict interactions and anticipate future states based on both 3D location and orientation, pushing the boundaries of spatial reasoning to include dynamic scenarios.

%% file: sec/appendix/1_examples.tex
\section{Examples of Benchmarks}

In this section, we present additional examples from the \dataset{} benchmark. Each example consists of an image accompanied by seven questions spanning five levels of difficulty, as shown in \cref{fig:appendix}.

%% file: sec/appendix/2_prompts.tex
\section{Prompts for LLM}

In this section, we present the detailed prompts designed for evaluating the Large Language Models (LLMs) on the \dataset{} benchmark. Each level targets a specific skill set, ranging from identifying single objects to reasoning about spatial relationships and occlusions in 3D space.

The prompts are structured with clear instructions, encouraging the model to first describe the scene comprehensively before formulating the answer. This approach ensures that the reasoning process is explicit and easy for interpretable. The output format is standardized as a JSON object, enabling straightforward evaluation and comparison against expected results. Figures~\ref{fig:prompts_level1}--\ref{fig:prompts_level5-6d} provide detailed examples of the prompts used at each level:



\begin{figure}[H]
  \centering
   \includegraphics[width=0.9\linewidth]{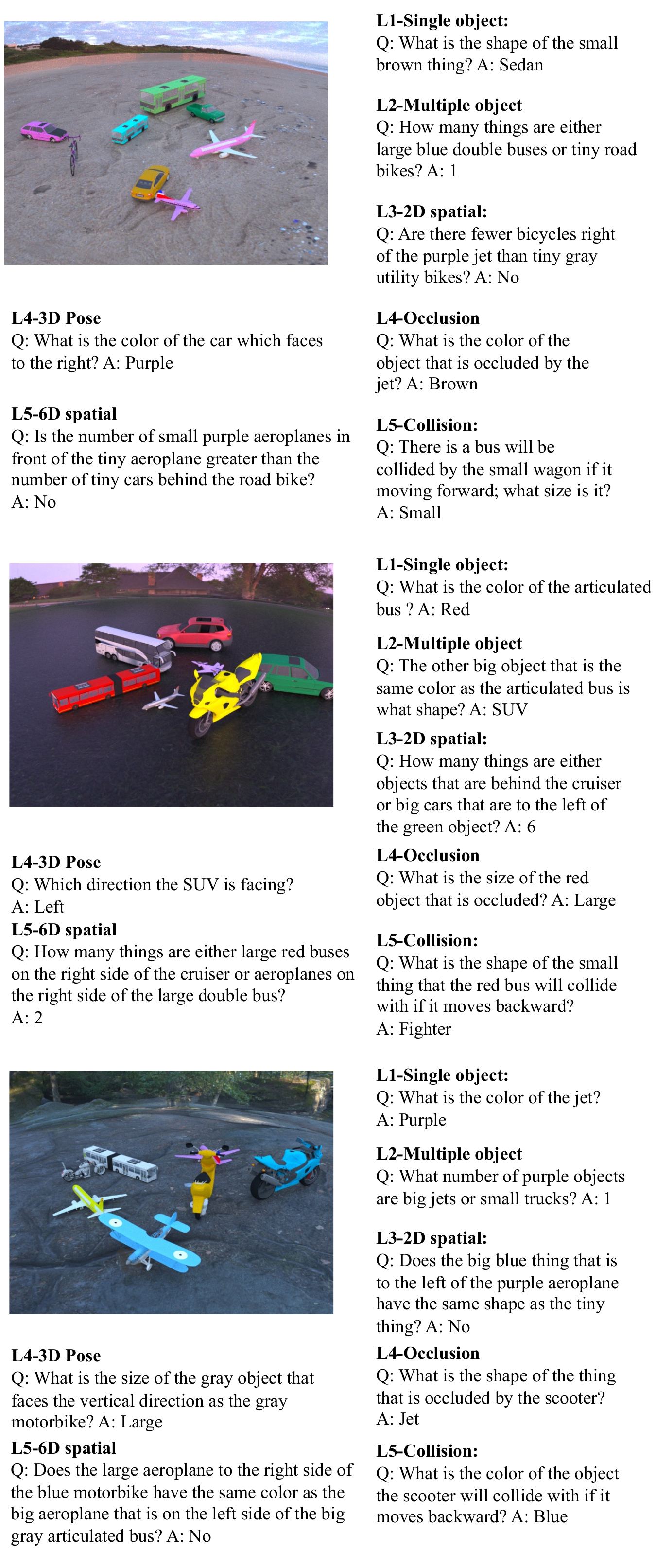}
   \caption{\textbf{Examples from the \dataset{} Benchmark.} Each example includes an image and seven questions that test the model across five difficulty levels.}
   \label{fig:appendix}
\end{figure}

\begin{figure*}[h]
    \centering
    \hypbox{Prompts for Level 1 - Single Object questions}{%
    \textcolor{ForestGreen}{\textbf{System:}} \\
    You are an intelligent chatbot designed to answer questions based on an image. 
    Your task is to analyze the images, identify attributes of the objects and then determine the answer to the question.\\

    INSTRUCTIONS:\\
    Step 1: Identify the objects mentioned in the question, noting each object's shape, color, size and their locations. Objects can have a shape (e.g., 'airliner'), size ('small' or 'large'), color (e.g., 'blue'). The possible colors are: gray, blue, purple, brown, green, cyan, red, and yellow. 

    Step 2: Use the identified details to form a brief reasoning process for answering the question. Based on this reasoning, provide your answer in one of the following formats: (1) a phrase chosen from the available options: Small, Large, Airliner, Dirtbike, Road bike, Tandem bike, Suv, Wagon, Scooter, Mountain bike, Minivan, Sedan, School bus, Fighter, Chopper, Double bus, Truck, Articulated bus, Cruiser, Jet, Utility bike, Regular bus, Biplane, Gray, Blue, Purple, Brown, Green, Cyan, Red, Yellow, Left, Right, Front, Back, (2) an integer between 0-10 for questions like 'How many' or 'What is the number of,' or (3) 'Yes' or 'No' for questions asking 'Is there.'
    
    Output your response using this JSON format:
    \{ 'Reasoning': '\textless reasoning\textgreater', 'Answer': '\textless answer\textgreater'\}
    \\

    \textcolor{RoyalBlue}{\textbf{User:}} \\
    What is the shape of the big yellow thing?\\

    \textcolor{Orange}{\textbf{GPT-4o:}} \\
    \{'Reasoning': 'The image contains a green utility bike with a red seat, a small white biplane, a large yellow sedan, and two purple tandem bikes. The large yellow object is a sedan.', 'Answer': 'Sedan'\}
    }
    \caption{\textbf{The prompts used for L1- Single object questions.} In the instruction, we encourage the model to first describe all the objects in the scenes and the reasoning process of answer the question. We format the reply into a dictionary format, where the answer can be decoded and compared with the ground truth answer.}
    \label{fig:prompts_level1}
\end{figure*}

\begin{figure*}[h]
    \centering
    \hypbox{Prompts for Level 2 - Multiple Objects questions}{%
    \textcolor{ForestGreen}{\textbf{System:}} \\
    You are an intelligent chatbot designed to answer questions based on an image. 
    Your task is to analyze the images, \textbf{identify attributes of multiple objects} and then determine the answer to the question.\\

    INSTRUCTIONS:\\
    Step 1: Identify the objects mentioned in the question, noting each object's shape, color, size and their locations. Objects can have a shape (e.g., 'airliner'), size ('small' or 'large'), color (e.g., 'blue'). The possible colors are: gray, blue, purple, brown, green, cyan, red, and yellow. 

    Step 2: Use the identified details to form a brief reasoning process for answering the question. Based on this reasoning, provide your answer in one of the following formats: (1) a phrase chosen from the available options: Small, Large, Airliner, Dirtbike, Road bike, Tandem bike, Suv, Wagon, Scooter, Mountain bike, Minivan, Sedan, School bus, Fighter, Chopper, Double bus, Truck, Articulated bus, Cruiser, Jet, Utility bike, Regular bus, Biplane, Gray, Blue, Purple, Brown, Green, Cyan, Red, Yellow, Left, Right, Front, Back, (2) an integer between 0-10 for questions like 'How many' or 'What is the number of,' or (3) 'Yes' or 'No' for questions asking 'Is there.'
    
    Output your response using this JSON format:
    { 'Reasoning': '\textless reasoning\textgreater', 'Answer': '\textless answer\textgreater', 'Related Objects': [{'shape': '\textless shape\textgreater', 'size': '\textless size\textgreater', 'color': '\textless color\textgreater'}, ...] }\\

    \textcolor{RoyalBlue}{\textbf{User:}} \\
    Is there any other thing that is the same color as the small utility bike?\\
    
    \textcolor{Orange}{\textbf{GPT-4o:}} \\
    \{'Reasoning': 'The objects in the image are: a small gray SUV, a small gray sedan, a small gray school bus, a small red utility bike, a small red tandem bike, a small red road bike, a small red mountain bike, a small red fighter jet, and a small purple biplane. The small utility bike is red. There are other objects that are also red: the tandem bike, the road bike, the mountain bike, and the fighter jet.', 'Answer': 'Yes'\}
    }
    \caption{\textbf{The prompts used for L2- Multiple object questions.} In the instruction, we encourage the model to first describe all the objects in the scenes and the reasoning process of answer the question. We format the reply into a dictionary format, where the answer can be decoded and compared with the ground truth answer.}
    \label{fig:prompts_level2}
\end{figure*}

\begin{figure*}[h]
    \centering
    \hypbox{Prompts for Level 3 - 2D Spatial questions}{%
    \textcolor{ForestGreen}{\textbf{System:}} \\
    You are an intelligent chatbot designed to answer questions based on an image. 
    Your task is to analyze the images, \textbf{identify attributes of multiple objects and their spatial relationship from 2D projected camera view}, and then determine the answer to the question.\\

    INSTRUCTIONS:\\
    Step 1: Identify the objects mentioned in the question, noting each object's shape, color, size and their locations. Objects can have a shape (e.g., 'airliner'), size ('small' or 'large'), color (e.g., 'blue'). The possible colors are: gray, blue, purple, brown, green, cyan, red, and yellow. 

    Step 2: Use the identified details to form a brief reasoning process for answering the question. Based on this reasoning, provide your answer in one of the following formats: (1) a phrase chosen from the available options: Small, Large, Airliner, Dirtbike, Road bike, Tandem bike, Suv, Wagon, Scooter, Mountain bike, Minivan, Sedan, School bus, Fighter, Chopper, Double bus, Truck, Articulated bus, Cruiser, Jet, Utility bike, Regular bus, Biplane, Gray, Blue, Purple, Brown, Green, Cyan, Red, Yellow, Left, Right, Front, Back, (2) an integer between 0-10 for questions like 'How many' or 'What is the number of,' or (3) 'Yes' or 'No' for questions asking 'Is there.'
    
    Output your response using this JSON format:
    { 'Reasoning': '\textless reasoning\textgreater', 'Answer': '\textless answer\textgreater', 'Related Objects': [{'shape': '\textless shape\textgreater', 'size': '\textless size\textgreater', 'color': '\textless color\textgreater'}, ...] }\\

    \textcolor{RoyalBlue}{\textbf{User:}} \\
    Is there a gray mountain bike that is to the left of the gray thing that is behind the tiny chopper?
    \\

    \textcolor{Orange}{\textbf{GPT-4o:}} \\
    \{'Reasoning': 'In the image, there are several objects including a gray car, a red car, a white bus, a purple airplane, and several bicycles. The gray car is located towards the left side of the image. To the right of the gray car, there is a red car. There are no brown bicycles to the right of the red car.', 'Answer': 'No'\}
    }
    \caption{\textbf{The prompts used for L3- 2D Spatial questions.} In the instruction, we encourage the model to first describe all the objects in the scenes and the reasoning process of answer the question. We format the reply into a dictionary format, where the answer can be decoded and compared with the ground truth answer.}
    \label{fig:prompts_level3}
\end{figure*}

\begin{figure*}[h]
    \centering
    \hypbox{Prompts for Level 4 - Occlusion questions}{%
    \textcolor{ForestGreen}{\textbf{System:}} \\
    You are an intelligent chatbot designed to answer questions based on an image. 
    Your task is to analyze the images, \textbf{identify attributes of multiple objects and their occlusions relationships}, and then determine the answer to the question.\\

    INSTRUCTIONS:\\
    Step 1: Identify the objects mentioned in the question, noting each object's shape, color, size and their locations. Objects can have a shape (e.g., 'airliner'), size ('small' or 'large'), color (e.g., 'blue'). The possible colors are: gray, blue, purple, brown, green, cyan, red, and yellow. 

    Step 2: Use the identified details to form a brief reasoning process for answering the question. Based on this reasoning, provide your answer in one of the following formats: (1) a phrase chosen from the available options: Small, Large, Airliner, Dirtbike, Road bike, Tandem bike, Suv, Wagon, Scooter, Mountain bike, Minivan, Sedan, School bus, Fighter, Chopper, Double bus, Truck, Articulated bus, Cruiser, Jet, Utility bike, Regular bus, Biplane, Gray, Blue, Purple, Brown, Green, Cyan, Red, Yellow, Left, Right, Front, Back, (2) an integer between 0-10 for questions like 'How many' or 'What is the number of,' or (3) 'Yes' or 'No' for questions asking 'Is there.'
    
    Output your response using this JSON format:
    { 'Reasoning': '\textless reasoning\textgreater', 'Answer': '\textless answer\textgreater', 'Related Objects': [{'shape': '\textless shape\textgreater', 'size': '\textless size\textgreater', 'color': '\textless color\textgreater'}, ...] }\\

    \textcolor{RoyalBlue}{\textbf{User:}} \\
    What is the size of the object that is occluded by the truck?\\

    \textcolor{Orange}{\textbf{GPT-4o:}} \\
    \{'Reasoning': 'The image contains several objects: a gray sedan, a white school bus, a red tandem bike, a red road bike, a red mountain bike, a purple jet, a gray wagon, and a red cruiser. The truck is occluding a small red road bike.', 'Answer': 'Small'\}
    }
    \caption{\textbf{The prompts used for L4- Occlusion questions.} In the instruction, we encourage the model to first describe all the objects in the scenes and the reasoning process of answer the question. We format the reply into a dictionary format, where the answer can be decoded and compared with the ground truth answer.}
    \label{fig:prompts_level4-occ}
\end{figure*}

\begin{figure*}[h]
    \centering
    \hypbox{Prompts for Level 4 - 3D Pose questions}{%
    \textcolor{ForestGreen}{\textbf{System:}} \\
    You are an intelligent chatbot designed to answer questions based on an image. 
    Your task is to analyze the images, \textbf{identify attributes of multiple objects and their facing direction in 3D space from the camera view}, and then determine the answer to the question.\\

    INSTRUCTIONS:\\
    Step 1: Identify the objects mentioned in the question, noting each object's shape, color, size, and direction which facing at. Objects can have a shape (e.g., 'airliner'), size ('small' or 'large'), color (e.g., 'blue'), \textbf{and direction ('left', 'right', 'front', 'back') in camera view.} The possible colors are: gray, blue, purple, brown, green, cyan, red, and yellow. 
    
    Step 2: Use the identified details to form a brief reasoning process for answering the question. Based on this reasoning, provide your answer in one of the following formats: (1) a phrase chosen from the available options: Small, Large, Airliner, Dirtbike, Road bike, Tandem bike, Suv, Wagon, Scooter, Mountain bike, Minivan, Sedan, School bus, Fighter, Chopper, Double bus, Truck, Articulated bus, Cruiser, Jet, Utility bike, Regular bus, Biplane, Gray, Blue, Purple, Brown, Green, Cyan, Red, Yellow, Left, Right, Front, Back, (2) an integer between 0-10 for questions like 'How many' or 'What is the number of,' or (3) 'Yes' or 'No' for questions asking 'Is there.'
    
    Output your response using this JSON format:
    { 'Reasoning': '\textless reasoning\textgreater', 'Answer': '\textless answer\textgreater', 'Related Objects': [{'shape': '\textless shape\textgreater', 'size': '\textless size\textgreater', 'color': '\textless color\textgreater', \textbf{'direction': '\textless direction\textgreater'}}, ...] }\\

    \textcolor{RoyalBlue}{\textbf{User:}} \\
    Which direction the suv is facing? \\

    \textcolor{Orange}{\textbf{GPT-4o:}} \\
    {'Reasoning': 'The image contains several objects: a small gray SUV, a small gray wagon, a small white school bus, a small red sedan, a small purple jet, a small red tandem bike, a small red road bike, a small red mountain bike, and a small red utility bike. The SUV is positioned towards the left side of the image.', 'Answer': 'Left'}
    }
    \caption{\textbf{The prompts used for Level 4 - 3D Pose  questions.} In the instruction, we encourage the model to first describe all the objects in the scenes and the reasoning process of answer the question. We format the reply into a dictionary format, where the answer can be decoded and compared with the ground truth answer.}
    \label{fig:prompts_level-pose}
\end{figure*}

\begin{figure*}[h]
    \centering
    \hypbox{Prompts for Level 5 - Collision questions}{%
    \textcolor{ForestGreen}{\textbf{System:}} \\
    You are an intelligent chatbot designed to answer questions based on an image. 
    Your task is to analyze the images, \textbf{identify attributes of multiple objects and their potential collision given the assumption of moving direction in 3D space}, and then determine the answer to the question.\\

    INSTRUCTIONS:\\
    Step 1: Identify the objects mentioned in the question, noting each object's shape, color, size, and direction which facing at. Objects can have a shape (e.g., 'airliner'), size ('small' or 'large'), color (e.g., 'blue'), \textbf{and direction ('left', 'right', 'front', 'back') in camera view.} The possible colors are: gray, blue, purple, brown, green, cyan, red, and yellow. 
    
    Step 2: Use the identified details to form a brief reasoning process for answering the question. Based on this reasoning, provide your answer in one of the following formats: (1) a phrase chosen from the available options: Small, Large, Airliner, Dirtbike, Road bike, Tandem bike, Suv, Wagon, Scooter, Mountain bike, Minivan, Sedan, School bus, Fighter, Chopper, Double bus, Truck, Articulated bus, Cruiser, Jet, Utility bike, Regular bus, Biplane, Gray, Blue, Purple, Brown, Green, Cyan, Red, Yellow, Left, Right, Front, Back, (2) an integer between 0-10 for questions like 'How many' or 'What is the number of,' or (3) 'Yes' or 'No' for questions asking 'Is there.'
    
    Output your response using this JSON format:
    { 'Reasoning': '\textless reasoning\textgreater', 'Answer': '\textless answer\textgreater', 'Related Objects': [{'shape': '\textless shape\textgreater', 'size': '\textless size\textgreater', 'color': '\textless color\textgreater', \textbf{'direction': '\textless direction\textgreater'}}, ...] }\\

    \textcolor{RoyalBlue}{\textbf{User:}} \\
    There is a small bicycle that is behind the aeroplane; how many tiny purple fighters are on the right side of it? \\

    \textcolor{Orange}{\textbf{GPT-4o:}} \\
    \{'Reasoning': 'In the image, there is a small bicycle behind the airplane. There is one small purple fighter on the right side of the small bicycle.', 'Answer': '1'\}
    }
    \caption{\textbf{The prompts used for Level 5 - Collision questions.} In the instruction, we encourage the model to first describe all the objects in the scenes and the reasoning process of answer the question. We format the reply into a dictionary format, where the answer can be decoded and compared with the ground truth answer.}
    \label{fig:prompts_level-col}
\end{figure*}

\begin{figure*}[h]
    \centering
    \hypbox{Prompts for Level 5 - 6D Spatial questions}{%
    \textcolor{ForestGreen}{\textbf{System:}} \\
    You are an intelligent chatbot designed to answer questions based on an image. 
    Your task is to analyze the images, \textbf{identify attributes of multiple objects and their spatial relationship from objects' perspective in 3D space}, and then determine the answer to the question.\\

    INSTRUCTIONS:\\
    Step 1: Identify the objects mentioned in the question, noting each object's shape, color, size, and direction which facing at. Objects can have a shape (e.g., 'airliner'), size ('small' or 'large'), color (e.g., 'blue'), \textbf{and direction ('left', 'right', 'front', 'back') in camera view.} The possible colors are: gray, blue, purple, brown, green, cyan, red, and yellow. 
    
    Step 2: Use the identified details to form a brief reasoning process for answering the question. Based on this reasoning, provide your answer in one of the following formats: (1) a phrase chosen from the available options: Small, Large, Airliner, Dirtbike, Road bike, Tandem bike, Suv, Wagon, Scooter, Mountain bike, Minivan, Sedan, School bus, Fighter, Chopper, Double bus, Truck, Articulated bus, Cruiser, Jet, Utility bike, Regular bus, Biplane, Gray, Blue, Purple, Brown, Green, Cyan, Red, Yellow, Left, Right, Front, Back, (2) an integer between 0-10 for questions like 'How many' or 'What is the number of,' or (3) 'Yes' or 'No' for questions asking 'Is there.'
    
    Output your response using this JSON format:
    { 'Reasoning': '\textless reasoning\textgreater', 'Answer': '\textless answer\textgreater', 'Related Objects': [{'shape': '\textless shape\textgreater', 'size': '\textless size\textgreater', 'color': '\textless color\textgreater', \textbf{'direction': '\textless direction\textgreater'}}, ...] }\\

    \textcolor{RoyalBlue}{\textbf{User:}} \\
    There is a aeroplane; how many tiny brown bicycles will it collide if it moving forward? \\

    \textcolor{Orange}{\textbf{GPT-4o:}} \\
    \{'Reasoning': 'The image contains a small purple jet, a small gray sedan, a small gray wagon, a small white school bus, a small red tandem bike, a small red road bike, a small brown road bike, a small brown mountain bike, and a small red sedan. The small purple jet is positioned at the back of the image. If it moves forward, it will collide with the small brown road bike and the small brown mountain bike.', 'Answer': '2'\}
    }
    \caption{\textbf{The prompts used for Level 5 - 6D Spatial questions.} In the instruction, we encourage the model to first describe all the objects in the scenes and the reasoning process of answer the question. We format the reply into a dictionary format, where the answer can be decoded and compared with the ground truth answer.}
    \label{fig:prompts_level5-6d}
\end{figure*}


%% file: sec/appendix/4_vis.tex
\section{Extension to Real Image Setting}
As described in \cref{sec:real}, we extend the \textit{L4-Pose} questions to the real-world image dataset SUN-RGBD~\cite{song2015sun}. The scene structure from SUN-RGBD~\cite{song2015sun} provides object categories and 3D bounding boxes. We calculate the rotation of bounding boxes along the z-axis as ground truth orientations and construct questions using the object category and orientation, naming the set \textit{L4-Pose-Real}. The question generation template remains the same as that of the original \textit{L4-Pose} questions.

To reduce ambiguity in object localization, we add a red bounding box around target objects in the images and reference them in the questions to ensure clarity in object identification within the scene. For example: "Which direction are the chairs in the red bounding box facing?"

%% file: main.bbl
\begin{thebibliography}{53}
\providecommand{\natexlab}[1]{#1}
\providecommand{\url}[1]{\texttt{#1}}
\expandafter\ifx\csname urlstyle\endcsname\relax
  \providecommand{\doi}[1]{doi: #1}\else
  \providecommand{\doi}{doi: \begingroup \urlstyle{rm}\Url}\fi

\bibitem[Academy(2023)]{alibaba2023gemini}
Alibaba~DAMO Academy.
\newblock Gemini: Vision-language model by alibaba.
\newblock \emph{Alibaba DAMO}, 2023.

\bibitem[An et~al.(2024)An, Geng, Chen, Li, Dou, and Dong]{an2024rgbmanip}
Boshi An, Yiran Geng, Kai Chen, Xiaoqi Li, Qi Dou, and Hao Dong.
\newblock Rgbmanip: Monocular image-based robotic manipulation through active object pose estimation.
\newblock In \emph{2024 IEEE International Conference on Robotics and Automation (ICRA)}, pages 7748--7755. IEEE, 2024.

\bibitem[Anthropic(2023)]{anthropic2023claude}
Anthropic.
\newblock Claude: An anthropic language model.
\newblock \emph{Anthropic}, 2023.

\bibitem[Baruch et~al.(2021)Baruch, Chen, Dehghan, Dimry, Feigin, Fu, Gebauer, Joffe, Kurz, Schwartz, et~al.]{baruch2021arkitscenes}
Gilad Baruch, Zhuoyuan Chen, Afshin Dehghan, Tal Dimry, Yuri Feigin, Peter Fu, Thomas Gebauer, Brandon Joffe, Daniel Kurz, Arik Schwartz, et~al.
\newblock Arkitscenes: A diverse real-world dataset for 3d indoor scene understanding using mobile rgb-d data.
\newblock \emph{arXiv preprint arXiv:2111.08897}, 2021.

\bibitem[Billinghurst et~al.(2015)Billinghurst, Clark, and Lee]{billinghurst2015survey}
Mark Billinghurst, Adrian Clark, and Gun Lee.
\newblock A survey of augmented reality.
\newblock \emph{Foundations and Trends in Human–Computer Interaction}, 8\penalty0 (2-3):\penalty0 73--272, 2015.

\bibitem[Brazil et~al.(2023)Brazil, Kumar, Straub, Ravi, Johnson, and Gkioxari]{brazil2023omni3d}
Garrick Brazil, Abhinav Kumar, Julian Straub, Nikhila Ravi, Justin Johnson, and Georgia Gkioxari.
\newblock Omni3d: A large benchmark and model for 3d object detection in the wild.
\newblock In \emph{Proceedings of the IEEE/CVF conference on computer vision and pattern recognition}, pages 13154--13164, 2023.

\bibitem[Caesar et~al.(2020)Caesar, Bankiti, Lang, Vora, Liong, Xu, Krishnan, Pan, Baldan, and Beijbom]{caesar2020nuscenes}
Holger Caesar, Varun Bankiti, Alex~H Lang, Sourabh Vora, Venice~Erin Liong, Qiang Xu, Anush Krishnan, Yu Pan, Giancarlo Baldan, and Oscar Beijbom.
\newblock nuscenes: A multimodal dataset for autonomous driving.
\newblock In \emph{Proceedings of the IEEE/CVF conference on computer vision and pattern recognition}, pages 11621--11631, 2020.

\bibitem[Carion et~al.(2020)Carion, Massa, Synnaeve, Usunier, Kirillov, and Zagoruyko]{carion2020end}
Nicolas Carion, Francisco Massa, Gabriel Synnaeve, Nicolas Usunier, Alexander Kirillov, and Sergey Zagoruyko.
\newblock End-to-end object detection with transformers.
\newblock In \emph{European conference on computer vision}, pages 213--229. Springer, 2020.

\bibitem[Chen et~al.(2024)Chen, Xu, Kirmani, Ichter, Sadigh, Guibas, and Xia]{chen2024spatialvlm}
Boyuan Chen, Zhuo Xu, Sean Kirmani, Brain Ichter, Dorsa Sadigh, Leonidas Guibas, and Fei Xia.
\newblock Spatialvlm: Endowing vision-language models with spatial reasoning capabilities.
\newblock In \emph{Proceedings of the IEEE/CVF Conference on Computer Vision and Pattern Recognition}, pages 14455--14465, 2024.

\bibitem[Cheng et~al.(2024)Cheng, Yin, Fu, Guo, Yang, Kautz, Wang, and Liu]{cheng2024spatialrgpt}
An-Chieh Cheng, Hongxu Yin, Yang Fu, Qiushan Guo, Ruihan Yang, Jan Kautz, Xiaolong Wang, and Sifei Liu.
\newblock Spatialrgpt: Grounded spatial reasoning in vision language model.
\newblock \emph{arXiv preprint arXiv:2406.01584}, 2024.

\bibitem[Chiang et~al.(2023)Chiang, Li, Lin, Sheng, Wu, Zhang, Zheng, Zhuang, Zhuang, Gonzalez, et~al.]{chiang2023vicuna}
Wei-Lin Chiang, Zhuohan Li, Zi Lin, Ying Sheng, Zhanghao Wu, Hao Zhang, Lianmin Zheng, Siyuan Zhuang, Yonghao Zhuang, Joseph~E Gonzalez, et~al.
\newblock Vicuna: An open-source chatbot impressing gpt-4 with 90\%* chatgpt quality, march 2023.
\newblock \emph{URL https://lmsys. org/blog/2023-03-30-vicuna}, 3\penalty0 (5), 2023.

\bibitem[Duan et~al.(2024)Duan, Yang, Qiao, Fang, Chen, Liu, Dong, Zang, Zhang, Wang, et~al.]{duan2024vlmevalkit}
Haodong Duan, Junming Yang, Yuxuan Qiao, Xinyu Fang, Lin Chen, Yuan Liu, Xiaoyi Dong, Yuhang Zang, Pan Zhang, Jiaqi Wang, et~al.
\newblock Vlmevalkit: An open-source toolkit for evaluating large multi-modality models.
\newblock In \emph{Proceedings of the 32nd ACM international conference on multimedia}, pages 11198--11201, 2024.

\bibitem[Everingham et~al.(2010)Everingham, Van~Gool, Williams, Winn, and Zisserman]{everingham2010pascal}
Mark Everingham, Luc Van~Gool, Christopher~KI Williams, John Winn, and Andrew Zisserman.
\newblock The pascal visual object classes (voc) challenge.
\newblock \emph{International Journal of Computer Vision}, 88\penalty0 (2):\penalty0 303--338, 2010.

\bibitem[Geiger et~al.(2012)Geiger, Lenz, and Urtasun]{geiger2012we}
Andreas Geiger, Philip Lenz, and Raquel Urtasun.
\newblock Are we ready for autonomous driving? the kitti vision benchmark suite.
\newblock In \emph{Proceedings of the IEEE Conference on Computer Vision and Pattern Recognition (CVPR)}, pages 3354--3361, 2012.

\bibitem[Goyal et~al.(2017)Goyal, Khot, Summers-Stay, Batra, and Parikh]{goyal2017making}
Yash Goyal, Tejas Khot, Douglas Summers-Stay, Dhruv Batra, and Devi Parikh.
\newblock Making the v in vqa matter: Elevating the role of image understanding in visual question answering.
\newblock In \emph{Proceedings of the IEEE conference on computer vision and pattern recognition}, pages 6904--6913, 2017.

\bibitem[Hess(2013)]{hess2013blender}
Roland Hess.
\newblock \emph{Blender foundations: The essential guide to learning blender 2.5}.
\newblock Routledge, 2013.

\bibitem[Hu et~al.(2022)Hu, Hua, Yang, Shi, Smith, and Luo]{hu2022promptcap}
Yushi Hu, Hang Hua, Zhengyuan Yang, Weijia Shi, Noah~A Smith, and Jiebo Luo.
\newblock Promptcap: Prompt-guided task-aware image captioning.
\newblock \emph{arXiv preprint arXiv:2211.09699}, 2022.

\bibitem[Hudson and Manning(2019)]{hudson2019gqa}
Drew~A Hudson and Christopher~D Manning.
\newblock Gqa: A new dataset for real-world visual reasoning and compositional question answering.
\newblock In \emph{Proceedings of the IEEE/CVF conference on computer vision and pattern recognition}, pages 6700--6709, 2019.

\bibitem[Johnson et~al.(2017)Johnson, Hariharan, Van Der~Maaten, Fei-Fei, Lawrence~Zitnick, and Girshick]{johnson2017clevr}
Justin Johnson, Bharath Hariharan, Laurens Van Der~Maaten, Li Fei-Fei, C Lawrence~Zitnick, and Ross Girshick.
\newblock Clevr: A diagnostic dataset for compositional language and elementary visual reasoning.
\newblock In \emph{Proceedings of the IEEE conference on computer vision and pattern recognition}, pages 2901--2910, 2017.

\bibitem[Kamath et~al.(2021)Kamath, Singh, LeCun, Synnaeve, Misra, and Carion]{kamath2021mdetr}
Aishwarya Kamath, Mannat Singh, Yann LeCun, Gabriel Synnaeve, Ishan Misra, and Nicolas Carion.
\newblock Mdetr-modulated detection for end-to-end multi-modal understanding.
\newblock In \emph{Proceedings of the IEEE/CVF international conference on computer vision}, pages 1780--1790, 2021.

\bibitem[Kim et~al.(2019)Kim, Choi, Oh, and Kweon]{kim2019dense}
Dong-Jin Kim, Jinsoo Choi, Tae-Hyun Oh, and In~So Kweon.
\newblock Dense relational captioning: Triple-stream networks for relationship-based captioning.
\newblock In \emph{Proceedings of the IEEE/CVF Conference on Computer Vision and Pattern Recognition}, pages 6271--6280, 2019.

\bibitem[Kuznetsova et~al.(2020)Kuznetsova, Rom, Alldrin, Uijlings, Krasin, Pont-Tuset, Muryshkin, Popov, Elhachem, de~Carvalho, et~al.]{kuznetsova2020open}
Alina Kuznetsova, Hassan Rom, Neil Alldrin, Jasper Uijlings, Ilya Krasin, Jordi Pont-Tuset, Gleb Muryshkin, Sergey Popov, Marwa Elhachem, Pedro de Carvalho, et~al.
\newblock The open images dataset v4: Unified image classification, object detection, and visual relationship detection at scale.
\newblock \emph{International Journal of Computer Vision}, 128\penalty0 (7):\penalty0 1956--1981, 2020.

\bibitem[Li et~al.(2023{\natexlab{a}})Li, Li, Savarese, and Hoi]{li2023blip}
Junnan Li, Dongxu Li, Silvio Savarese, and Steven Hoi.
\newblock Blip-2: Bootstrapping language-image pre-training with frozen image encoders and large language models.
\newblock In \emph{International conference on machine learning}, pages 19730--19742. PMLR, 2023{\natexlab{a}}.

\bibitem[Li et~al.(2023{\natexlab{b}})Li, Wang, Stengel-Eskin, Kortylewski, Ma, Van~Durme, and Yuille]{li2023super}
Zhuowan Li, Xingrui Wang, Elias Stengel-Eskin, Adam Kortylewski, Wufei Ma, Benjamin Van~Durme, and Alan~L Yuille.
\newblock Super-clevr: A virtual benchmark to diagnose domain robustness in visual reasoning.
\newblock In \emph{Proceedings of the IEEE/CVF Conference on Computer Vision and Pattern Recognition}, pages 14963--14973, 2023{\natexlab{b}}.

\bibitem[Lindstr{\"o}m and Abraham(2022)]{lindstrom2022clevr}
Adam~Dahlgren Lindstr{\"o}m and Savitha~Sam Abraham.
\newblock Clevr-math: A dataset for compositional language, visual and mathematical reasoning.
\newblock \emph{arXiv preprint arXiv:2208.05358}, 2022.

\bibitem[Liu et~al.(2023{\natexlab{a}})]{liu2023llava}
Haotian Liu et~al.
\newblock Llava: Large language and vision assistant.
\newblock \emph{LLaVA Project}, 2023{\natexlab{a}}.

\bibitem[Liu et~al.(2023{\natexlab{b}})]{liu2023llavanext}
Haotian Liu et~al.
\newblock Llava-next: Extended large language and vision assistant.
\newblock \emph{LLaVA Project}, 2023{\natexlab{b}}.

\bibitem[Mao et~al.(2019)Mao, Gan, Kohli, Tenenbaum, and Wu]{mao2019neuro}
Jiayuan Mao, Chuang Gan, Pushmeet Kohli, Joshua~B Tenenbaum, and Jiajun Wu.
\newblock The neuro-symbolic concept learner: Interpreting scenes, words, and sentences from natural supervision.
\newblock \emph{arXiv preprint arXiv:1904.12584}, 2019.

\bibitem[OpenAI(2023)]{openai2023gpt4}
OpenAI.
\newblock Gpt-4 technical report.
\newblock \emph{OpenAI}, 2023.

\bibitem[Qiu and Yuille(2016)]{qiu2016unrealcv}
Weichao Qiu and Alan Yuille.
\newblock Unrealcv: Connecting computer vision to unreal engine.
\newblock In \emph{Computer Vision--ECCV 2016 Workshops: Amsterdam, The Netherlands, October 8-10 and 15-16, 2016, Proceedings, Part III 14}, pages 909--916. Springer, 2016.

\bibitem[Radford et~al.(2021)Radford, Kim, Hallacy, Ramesh, Goh, Agarwal, Sastry, Askell, Mishkin, Clark, et~al.]{radford2021learning}
Alec Radford, Jong~Wook Kim, Chris Hallacy, Aditya Ramesh, Gabriel Goh, Sandhini Agarwal, Girish Sastry, Amanda Askell, Pamela Mishkin, Jack Clark, et~al.
\newblock Learning transferable visual models from natural language supervision.
\newblock In \emph{International conference on machine learning}, pages 8748--8763. PMLR, 2021.

\bibitem[Ray et~al.(2024)Ray, Duan, Tan, Bashkirova, Hendrix, Ehsani, Kembhavi, Plummer, Krishna, Zeng, et~al.]{ray2024sat}
Arijit Ray, Jiafei Duan, Reuben Tan, Dina Bashkirova, Rose Hendrix, Kiana Ehsani, Aniruddha Kembhavi, Bryan~A Plummer, Ranjay Krishna, Kuo-Hao Zeng, et~al.
\newblock Sat: Spatial aptitude training for multimodal language models.
\newblock \emph{arXiv preprint arXiv:2412.07755}, 2024.

\bibitem[Ren et~al.(2016)Ren, He, Girshick, and Sun]{ren2016faster}
Shaoqing Ren, Kaiming He, Ross Girshick, and Jian Sun.
\newblock Faster r-cnn: Towards real-time object detection with region proposal networks.
\newblock \emph{IEEE transactions on pattern analysis and machine intelligence}, 39\penalty0 (6):\penalty0 1137--1149, 2016.

\bibitem[Sanders(2016)]{sanders2016introduction}
Andrew Sanders.
\newblock \emph{An introduction to Unreal engine 4}.
\newblock AK Peters/CRC Press, 2016.

\bibitem[Singh et~al.(2019)Singh, Natarajan, Shah, Jiang, Chen, Batra, Parikh, and Rohrbach]{singh2019towards}
Amanpreet Singh, Vivek Natarajan, Meet Shah, Yu Jiang, Xinlei Chen, Dhruv Batra, Devi Parikh, and Marcus Rohrbach.
\newblock Towards vqa models that can read.
\newblock In \emph{Proceedings of the IEEE/CVF conference on computer vision and pattern recognition}, pages 8317--8326, 2019.

\bibitem[Song et~al.(2015)Song, Lichtenberg, and Xiao]{song2015sun}
Shuran Song, Samuel~P Lichtenberg, and Jianxiong Xiao.
\newblock Sun rgb-d: A rgb-d scene understanding benchmark suite.
\newblock In \emph{Proceedings of the IEEE conference on computer vision and pattern recognition}, pages 567--576, 2015.

\bibitem[Team et~al.(2024)Team, Georgiev, Lei, Burnell, Bai, Gulati, Tanzer, Vincent, Pan, Wang, et~al.]{team2024gemini}
Gemini Team, Petko Georgiev, Ving~Ian Lei, Ryan Burnell, Libin Bai, Anmol Gulati, Garrett Tanzer, Damien Vincent, Zhufeng Pan, Shibo Wang, et~al.
\newblock Gemini 1.5: Unlocking multimodal understanding across millions of tokens of context.
\newblock \emph{arXiv preprint arXiv:2403.05530}, 2024.

\bibitem[Team(2023{\natexlab{a}})]{intern2023vl2}
Intern Team.
\newblock Internvl 2.0: A vision-language model.
\newblock \emph{Intern Foundation}, 2023{\natexlab{a}}.

\bibitem[Team(2023{\natexlab{b}})]{qwen2023vl7b}
Qwen Team.
\newblock Qwen-vl: A large multimodal model.
\newblock \emph{Alibaba Group}, 2023{\natexlab{b}}.

\bibitem[Tong et~al.(2024)Tong, Brown, Wu, Woo, Middepogu, Akula, Yang, Yang, Iyer, Pan, et~al.]{tong2024cambrian}
Shengbang Tong, Ellis Brown, Penghao Wu, Sanghyun Woo, Manoj Middepogu, Sai~Charitha Akula, Jihan Yang, Shusheng Yang, Adithya Iyer, Xichen Pan, et~al.
\newblock Cambrian-1: A fully open, vision-centric exploration of multimodal llms.
\newblock \emph{arXiv preprint arXiv:2406.16860}, 2024.

\bibitem[Touvron et~al.(2023)]{touvron2023llama}
Hugo Touvron et~al.
\newblock Llama: Open and efficient foundation language models.
\newblock \emph{Meta AI}, 2023.

\bibitem[Vinyals et~al.(2015)Vinyals, Toshev, Bengio, and Erhan]{vinyals2015show}
Oriol Vinyals, Alexander Toshev, Samy Bengio, and Dumitru Erhan.
\newblock Show and tell: A neural image caption generator.
\newblock In \emph{Proceedings of the IEEE conference on computer vision and pattern recognition}, pages 3156--3164, 2015.

\bibitem[Wang et~al.(2019)Wang, Sridhar, Huang, Valentin, Song, and Guibas]{wang2019normalized}
He Wang, Srinath Sridhar, Jingwei Huang, Julien Valentin, Shuran Song, and Leonidas~J Guibas.
\newblock Normalized object coordinate space for category-level 6d object pose and size estimation.
\newblock In \emph{Proceedings of the IEEE/CVF conference on computer vision and pattern recognition}, pages 2642--2651, 2019.

\bibitem[Wang et~al.(2023)Wang, Ma, Li, Kortylewski, and Yuille]{wang2023daware}
Xingrui Wang, Wufei Ma, Zhuowan Li, Adam Kortylewski, and Alan Yuille.
\newblock 3d-aware visual question answering about parts, poses and occlusions.
\newblock In \emph{Thirty-seventh Conference on Neural Information Processing Systems}, 2023.

\bibitem[Wang et~al.(2024)Wang, Ma, Li, Kortylewski, and Yuille]{wang20243d}
Xingrui Wang, Wufei Ma, Zhuowan Li, Adam Kortylewski, and Alan~L Yuille.
\newblock 3d-aware visual question answering about parts, poses and occlusions.
\newblock \emph{Advances in Neural Information Processing Systems}, 36, 2024.

\bibitem[Xiang et~al.(2014)Xiang, Mottaghi, and Savarese]{xiang2014beyond}
Yu Xiang, Roozbeh Mottaghi, and Silvio Savarese.
\newblock Beyond pascal: A benchmark for 3d object detection in the wild.
\newblock In \emph{IEEE winter conference on applications of computer vision}, pages 75--82. IEEE, 2014.

\bibitem[Xu et~al.(2015)Xu, Ba, Kiros, Cho, Courville, Salakhudinov, Zemel, and Bengio]{xu2015show}
Kelvin Xu, Jimmy Ba, Ryan Kiros, Kyunghyun Cho, Aaron Courville, Ruslan Salakhudinov, Rich Zemel, and Yoshua Bengio.
\newblock Show, attend and tell: Neural image caption generation with visual attention.
\newblock In \emph{International conference on machine learning}, pages 2048--2057. PMLR, 2015.

\bibitem[Xu et~al.(2024)Xu, Zhao, Zhou, Lin, Ng, and Feng]{xu2024pllava}
Lin Xu, Yilin Zhao, Daquan Zhou, Zhijie Lin, See~Kiong Ng, and Jiashi Feng.
\newblock Pllava: Parameter-free llava extension from images to videos for video dense captioning.
\newblock \emph{arXiv preprint arXiv:2404.16994}, 2024.

\bibitem[Yi et~al.(2018)Yi, Wu, Gan, Torralba, Kohli, and Tenenbaum]{yi2018neural}
Kexin Yi, Jiajun Wu, Chuang Gan, Antonio Torralba, Pushmeet Kohli, and Josh Tenenbaum.
\newblock Neural-symbolic vqa: Disentangling reasoning from vision and language understanding.
\newblock \emph{Advances in neural information processing systems}, 31, 2018.

\bibitem[Yi et~al.(2019)Yi, Gan, Li, Kohli, Wu, Torralba, and Tenenbaum]{yi2019clevrer}
Kexin Yi, Chuang Gan, Yunzhu Li, Pushmeet Kohli, Jiajun Wu, Antonio Torralba, and Joshua~B Tenenbaum.
\newblock Clevrer: Collision events for video representation and reasoning.
\newblock \emph{arXiv preprint arXiv:1910.01442}, 2019.

\bibitem[Yue et~al.(2024)Yue, Ni, Zhang, Zheng, Liu, Zhang, Stevens, Jiang, Ren, Sun, et~al.]{yue2024mmmu}
Xiang Yue, Yuansheng Ni, Kai Zhang, Tianyu Zheng, Ruoqi Liu, Ge Zhang, Samuel Stevens, Dongfu Jiang, Weiming Ren, Yuxuan Sun, et~al.
\newblock Mmmu: A massive multi-discipline multimodal understanding and reasoning benchmark for expert agi.
\newblock In \emph{Proceedings of the IEEE/CVF Conference on Computer Vision and Pattern Recognition}, pages 9556--9567, 2024.

\bibitem[Yuille and Kersten(2006)]{yuille2006vision}
Alan Yuille and Daniel Kersten.
\newblock Vision as bayesian inference: analysis by synthesis?
\newblock \emph{Trends in cognitive sciences}, 10\penalty0 (7):\penalty0 301--308, 2006.

\bibitem[Zhang et~al.(2024)Zhang, Huang, Peng, Wu, Hu, Chen, Zhao, and Dong]{zhang2024omni6dpose}
Jiyao Zhang, Weiyao Huang, Bo Peng, Mingdong Wu, Fei Hu, Zijian Chen, Bo Zhao, and Hao Dong.
\newblock Omni6dpose: A benchmark and model for universal 6d object pose estimation and tracking.
\newblock In \emph{European Conference on Computer Vision}, pages 199--216. Springer, 2024.

\end{thebibliography}
